%% file: main.tex
\documentclass{article}

\PassOptionsToPackage{numbers}{natbib}
 \usepackage[dblblindworkshop,final]{neurips_2025}

\workshoptitle{Lock-LLM Workshop: Prevent Unauthorized Knowledge Use from Large Language Models}

\usepackage[utf8]{inputenc} %
\usepackage[T1]{fontenc}    %
\usepackage{hyperref}       %
\usepackage{url}            %
\usepackage{booktabs}       %
\usepackage{amsfonts}       %
\usepackage{nicefrac}       %
\usepackage{microtype}      %
\usepackage{xcolor}         %
\usepackage{algorithm} %
\usepackage{algorithmicx} \usepackage{algcompatible}
\usepackage{algpseudocode}
\usepackage{amsmath}
\usepackage{graphicx}
\usepackage{array}
\usepackage{booktabs}
\usepackage{makecell}
\usepackage{adjustbox}
\usepackage{siunitx}

\title{Text-to-Image Models Leave Identifiable Signatures: Implications for Leaderboard Security}

\author{%
  Ali Naseh\thanks{Correspondence to anaseh@cs.umass.edu}\, \textsuperscript{1} \quad
  Anshuman Suri\textsuperscript{2} \quad
  Yuefeng Peng\textsuperscript{1} \quad
  Harsh Chaudhari\textsuperscript{2} \\
  \textbf{Alina Oprea}\textsuperscript{2} \quad
  \textbf{Amir Houmansadr}\textsuperscript{1}
  \\
  \textsuperscript{1}University of Massachusetts Amherst \quad
  \textsuperscript{2}Northeastern University \\
}

\begin{document}

\maketitle

\begin{abstract}
Generative AI leaderboards are central to evaluating model capabilities, but remain vulnerable to manipulation.
Among key adversarial objectives is \emph{rank manipulation}, where an attacker must first deanonymize the models behind displayed outputs---a threat previously demonstrated and explored for large language models (LLMs).  
We show that this problem can be even more severe for text-to-image leaderboards, where deanonymization is markedly easier.  
Using over 150{,}000 generated images from 280 prompts and 19 diverse models spanning multiple organizations, architectures, and sizes, we demonstrate that simple real-time classification in CLIP embedding space identifies the generating model with high accuracy, even without prompt control or historical data.  
We further introduce a prompt-level separability metric and identify prompts that enable near-perfect deanonymization.  
Our results indicate that rank manipulation in text-to-image leaderboards is easier than previously recognized, underscoring the need for stronger defenses.
\end{abstract}

\input{sections/01_intro}

\input{sections/02_related_work}
\input{sections/03_methodology}
\input{sections/04_experiments}

\input{sections/05_conclusion}

\bibliographystyle{unsrtnat} 
\bibliography{main}

\clearpage
\appendix
\section{Setup Details}
\label{app:setup_details}
\input{tables/model_lists}

\section{Additional Results}
\label{app:additional_results}

\subsection{Details of AUC and TPR Computation}\label{sec:auc_details}

For each target model we use as the decision score the margin
$\mathbf{TargetSim} - \mathbf{BestOtherSim}$,  
where $\mathrm{TargetSim}$ is the cosine similarity between the test image embedding and the
centroid of the target model’s cluster, and $\mathrm{BestOtherSim}$ is the highest such similarity
across all non-target models.  
ROC curves are then computed from this score, and we report ROC–AUC as well as TPR at low FPR operating points (e.g., 2\% and 5\%).

\subsection{Detection Without Access to Other Models}
\label{app:hardest_one_vs_rest}
In our most restrictive setting, the adversary seeks to determine whether a given image was generated by its target model but has access only to that model’s own generations for the same prompt.  
To classify the given sample, for the corresponding prompt we use 30 generations from the target model to build a centroid and compute a similarity threshold based on the $80^{\text{th}}$ percentile of in-cluster distances.  
Concretely, let $\mathbf{c}$ denote the L2-normalized centroid of these embeddings, and let $s_i=\langle \mathbf{x}_i,\mathbf{c}\rangle$ represent cosine similarities of the target model’s own generations.  
We define the similarity threshold as
$\mathrm{SimThresh}=1-\mathrm{quantile}_{0.8}(1-s_i)$.  
Given a test image with embedding $\mathbf{z}$, we compute $\mathrm{TargetSim}=\langle \mathbf{z},\mathbf{c}\rangle$ and use the margin
$\mathrm{TargetSim}-\mathrm{SimThresh}$ as a continuous decision score.  
A sample is classified as generated by the target if this margin is non-negative.  
ROC curves are derived from these scores, and we report ROC–AUC and TPR at low FPR operating points (e.g., 2\% and 5\%).  
Table~\ref{tab:one_vs_rest_all_fixed_no_knowledge} summarizes the results.  
Although performance decreases due to the lack of information about other models, some models—most notably \texttt{SDXL~Turbo}—still reach $99\%$ top-1 accuracy, highlighting the strength of their model-specific signatures.

\begin{figure}[t]
    \centering
    \includegraphics[width=0.85\linewidth]{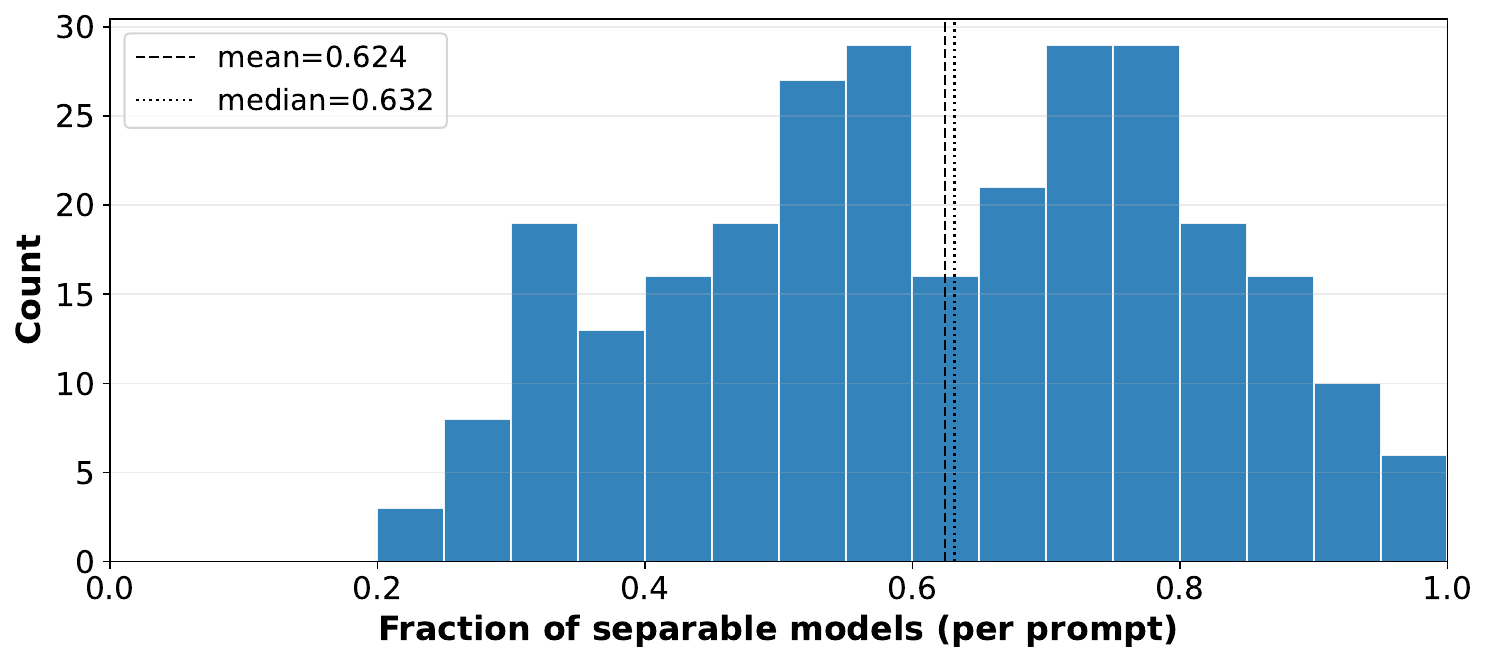}
    \caption{
        Distribution of the distinguishability score over the evaluation prompts.
    }
    \label{fig:nn_dist}
\end{figure}

\begin{figure}[t]
    \centering
    \includegraphics[width=0.85\linewidth]{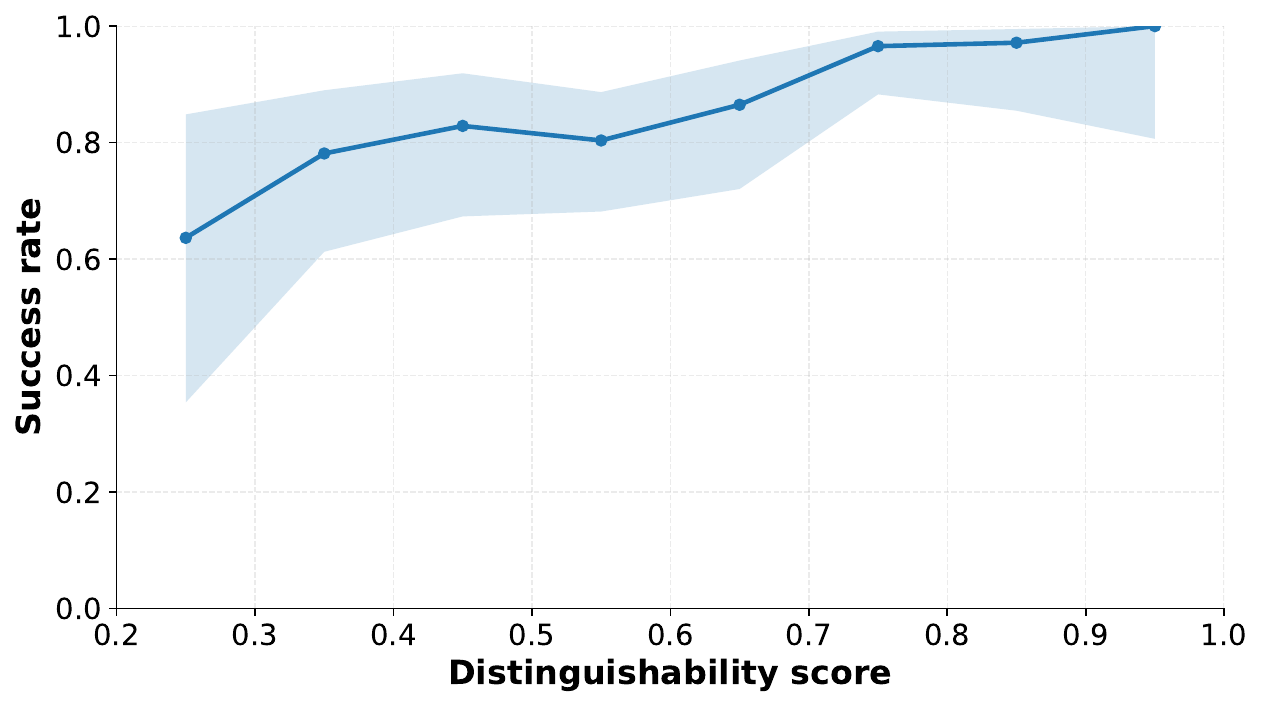}
    \caption{
        Relationship between prompt-level distinguishability and deanonymization accuracy.
        Each point represents an evaluation prompt, and the curve shows that higher distinguishability scores
        lead to consistently higher top-1 deanonymization accuracy.
        This confirms that the distinguishability metric is a strong predictor of attack success.
    }
    \label{fig:success_vs_dist}
\end{figure}

\input{tables/one_vs_rest_fixed_model}

\input{tables/one_vs_rest_fixed_model_no_knowledge}

\clearpage
\newpage

\end{document}

%% file: sections/01_intro.tex
\section{Introduction}

Generative AI leaderboards have become essential to the rapid progress and adoption of generative models, serving as public benchmarks that track and compare model capabilities. They provide standardized evaluations that guide research directions and inform deployment choices \cite{koch2024protoscience}, including dynamic query routing \cite{ong2025routellm,index_mixing}.
Broadly, leaderboards fall into two categories. Benchmark-based leaderboards rank models using predefined datasets and quantitative metrics, while voting-based leaderboards rely on user comparisons of model outputs to determine rankings.

Recent studies demonstrate how generative-model leaderboards are susceptible to various vulnerabilities such as \emph{rank manipulation} \cite{huang2025exploring,min2025improving} ---strategically biasing votes to promote or demote specific models.
A critical step in rank-manipulation attacks against leaderboards is \emph{model de\-anonymization}---identifying which models generated the content shown to voters. 
Prior works on LLM leaderboards assume that users can submit arbitrary prompts, or require access to historical prompt--response pairs to train deanonymization classifiers.
Realistically, however, leaderboards may restrict this freedom by providing the prompts themselves, making such attacks significantly harder.
We show that deanonymization can be \emph{easier} in text-to-image (T2I) leaderboards than text-based ones, even with no control over prompts, and without training any classifier.
We show that simple real-time embedding-space classification can accurately identify the underlying models.

We hypothesize that in T2I generation, the diversity of outputs from a given model across multiple generations of the same prompt is relatively low (Figure \ref{fig:prompt_diversity}). 
Moreover, these outputs often differ systematically from those of other models in terms of style, content, or other features not explicitly described in the prompt. 
Such differences naturally arise from variations in training data, architecture, and model size. 
This phenomenon causes generations from different models to form distinguishable clusters in the embedding space for most prompts, which adversaries can exploit for de\-anonymization.

To test this hypothesis, we analyze 280 prompts collected from a prominent T2I leaderboard and a diverse set of 19 T2I models spanning multiple organizations, architectures, and model sizes (both open source and commercial), producing over 150{,}000 images in total.  
We find that a straightforward real-time classification in the embedding space leads to high deanonymization accuracy.  
We further define a metric to quantify distinguishability between model generations per prompt, allowing us to identify prompts that yield complete separability in the embedding space.  
We find that such \emph{perfectly distinguishable} prompts exist and could be exploited if users were allowed to submit their own prompts. Deanonymization can also help amplify other attacks: once the generating T2I model is identified, an adversary can choose an appropriate surrogate and apply targeted prompt-optimization or iterative reproduction attacks to better replicate the original image \cite{naseh2024iteratively}.  
Together, these findings highlight the unique security threat posed by T2I models, particularly in voting-based leaderboards.

\input{figs/examples}

%% file: figs/examples.tex
\newcommand{\img}[1]{\raisebox{-0.5\height}{\includegraphics[height=2.2cm]{#1}}}

\begin{figure}[t]
\centering
\renewcommand{\arraystretch}{1.15}
\setlength{\tabcolsep}{3pt}

\begin{minipage}{0.91\linewidth}
\centering
\textbf{Prompt:} ``An impressionistic painting of a bustling city street in the rain, vibrant umbrellas dotting the crowd.''
\end{minipage}

\resizebox{\linewidth}{!}{%
\begin{tabular}{@{}>{\centering\arraybackslash}m{3.2cm}ccccc@{}}
\toprule
\textbf{Model} & \multicolumn{5}{c}{Generated Images (five different seeds)}\\
\midrule

\makecell[c]{Midjourney\\v6} &
\img{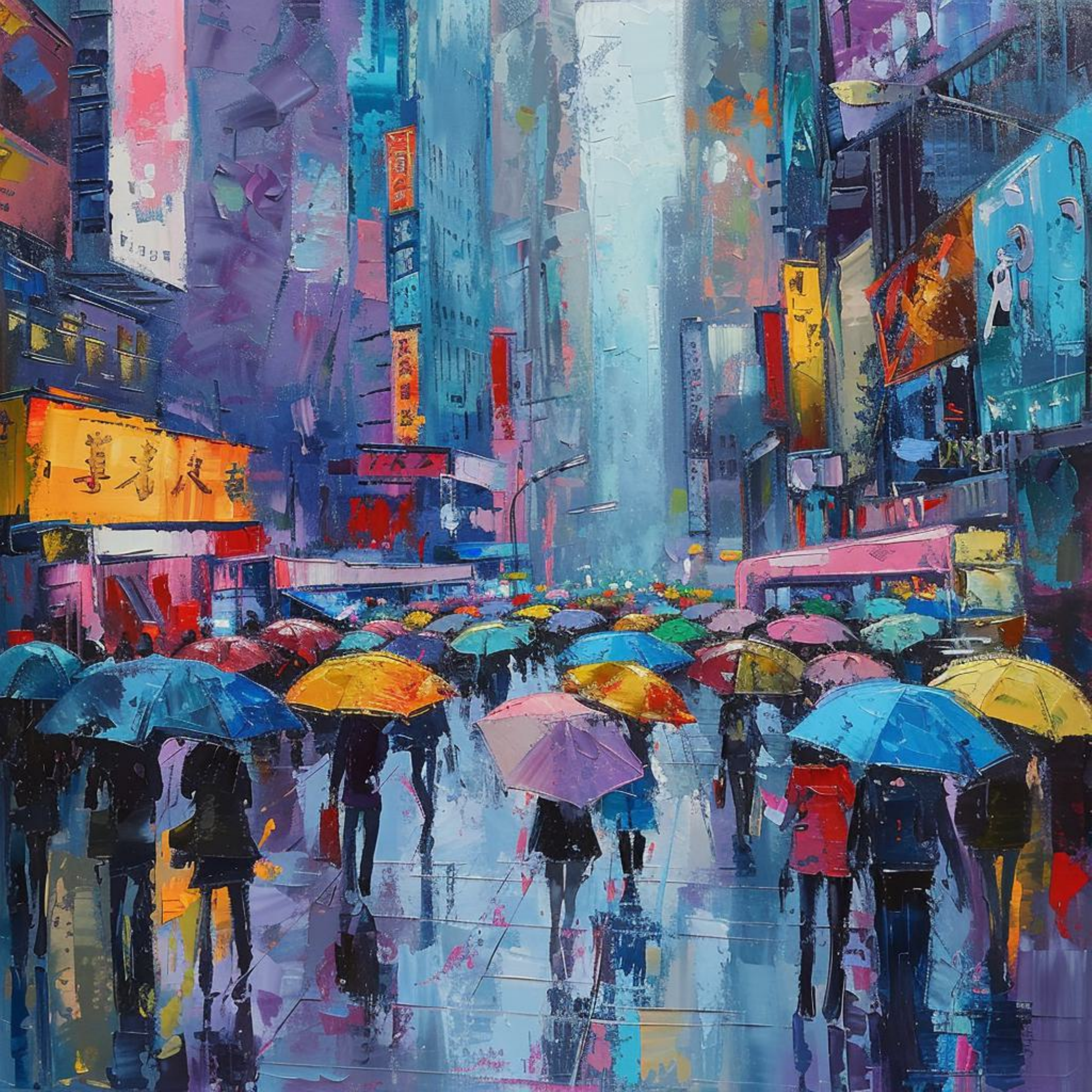} &
\img{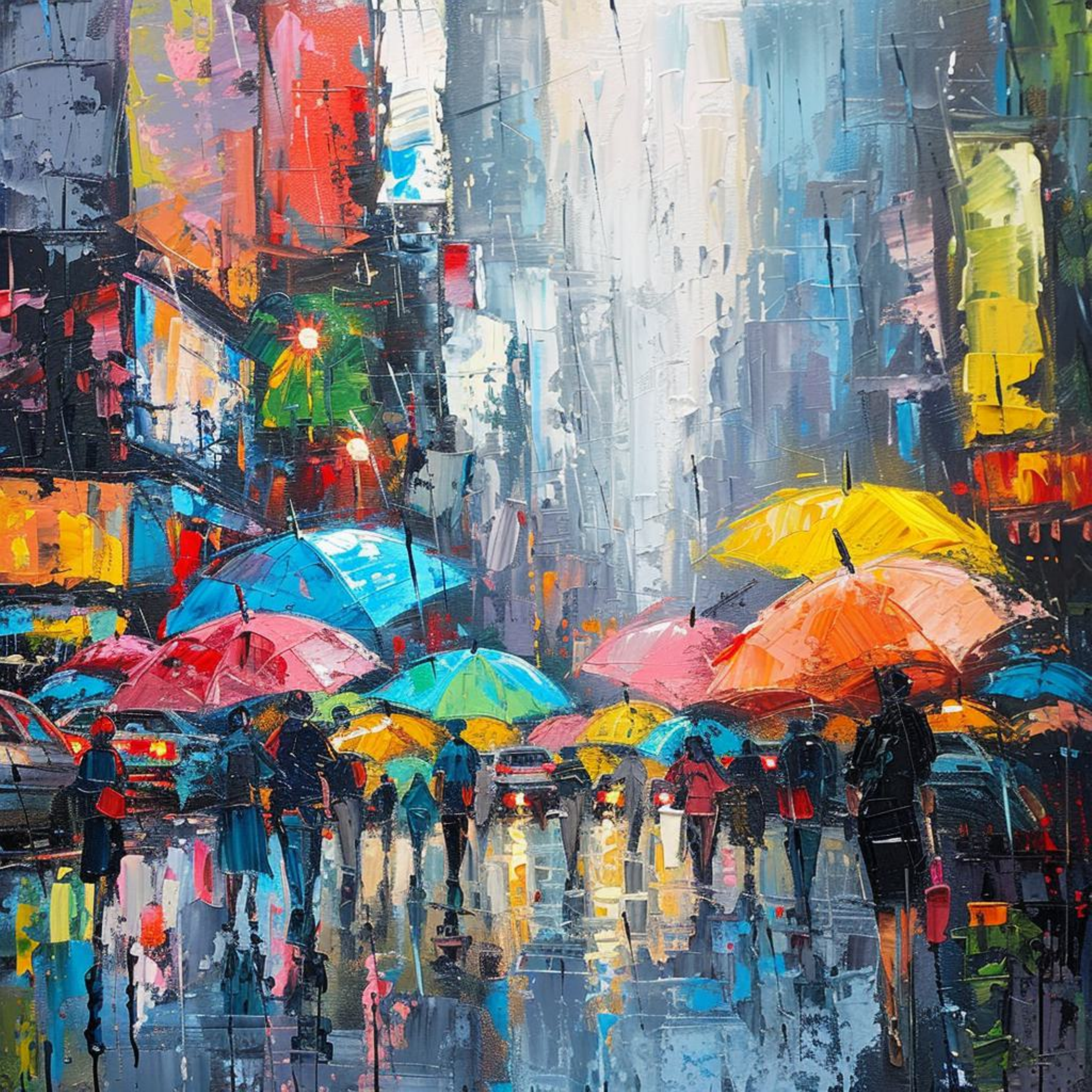} &
\img{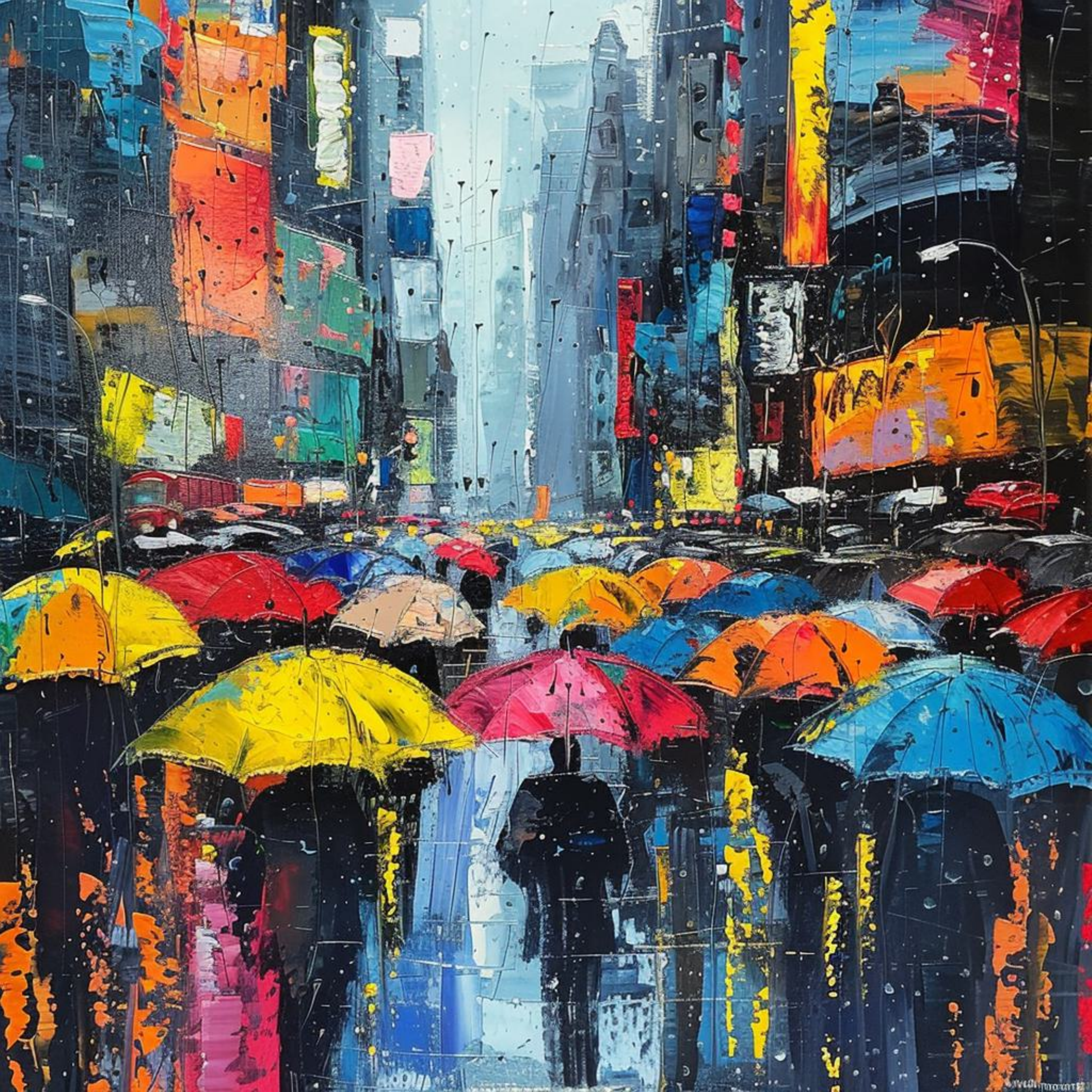} &
\img{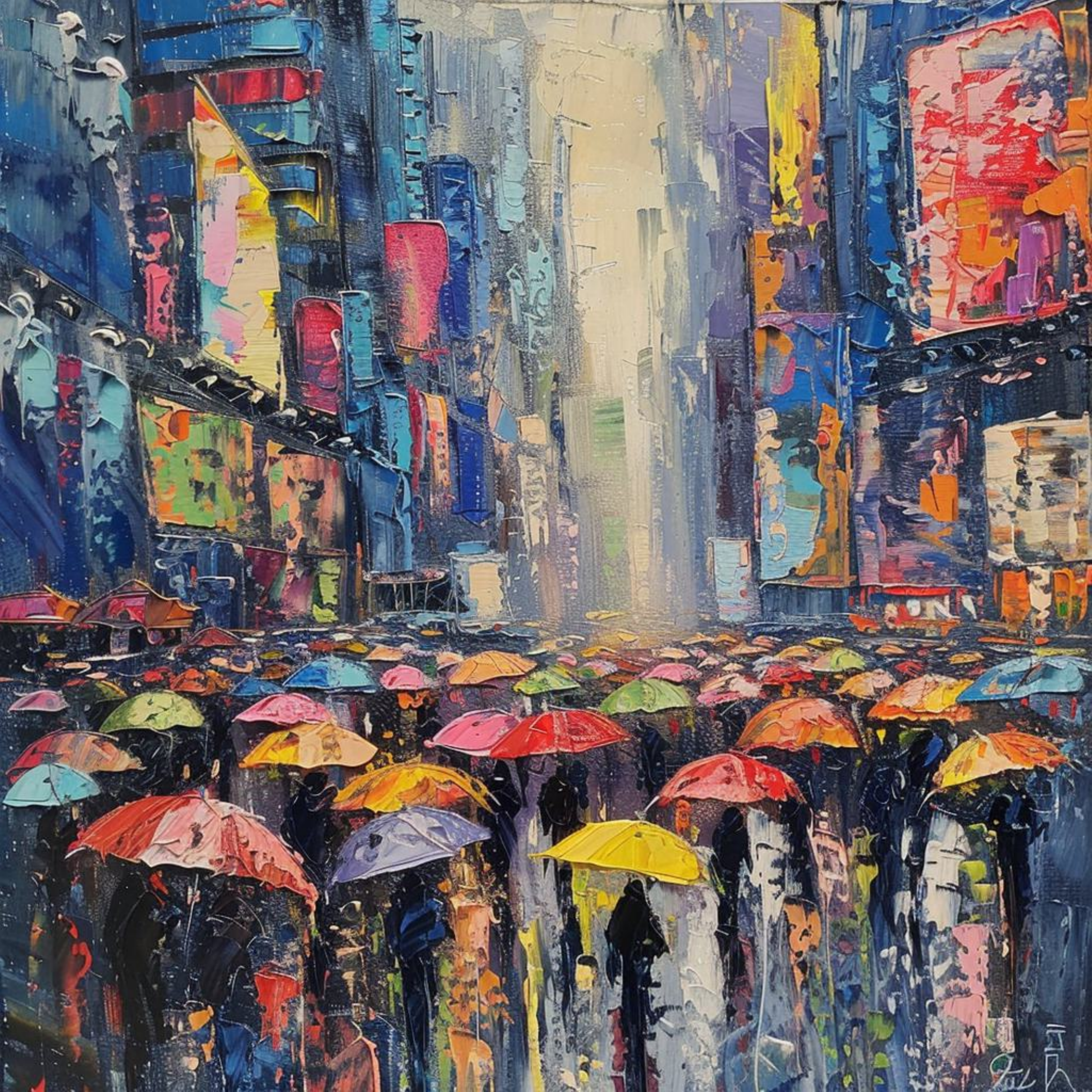} &
\img{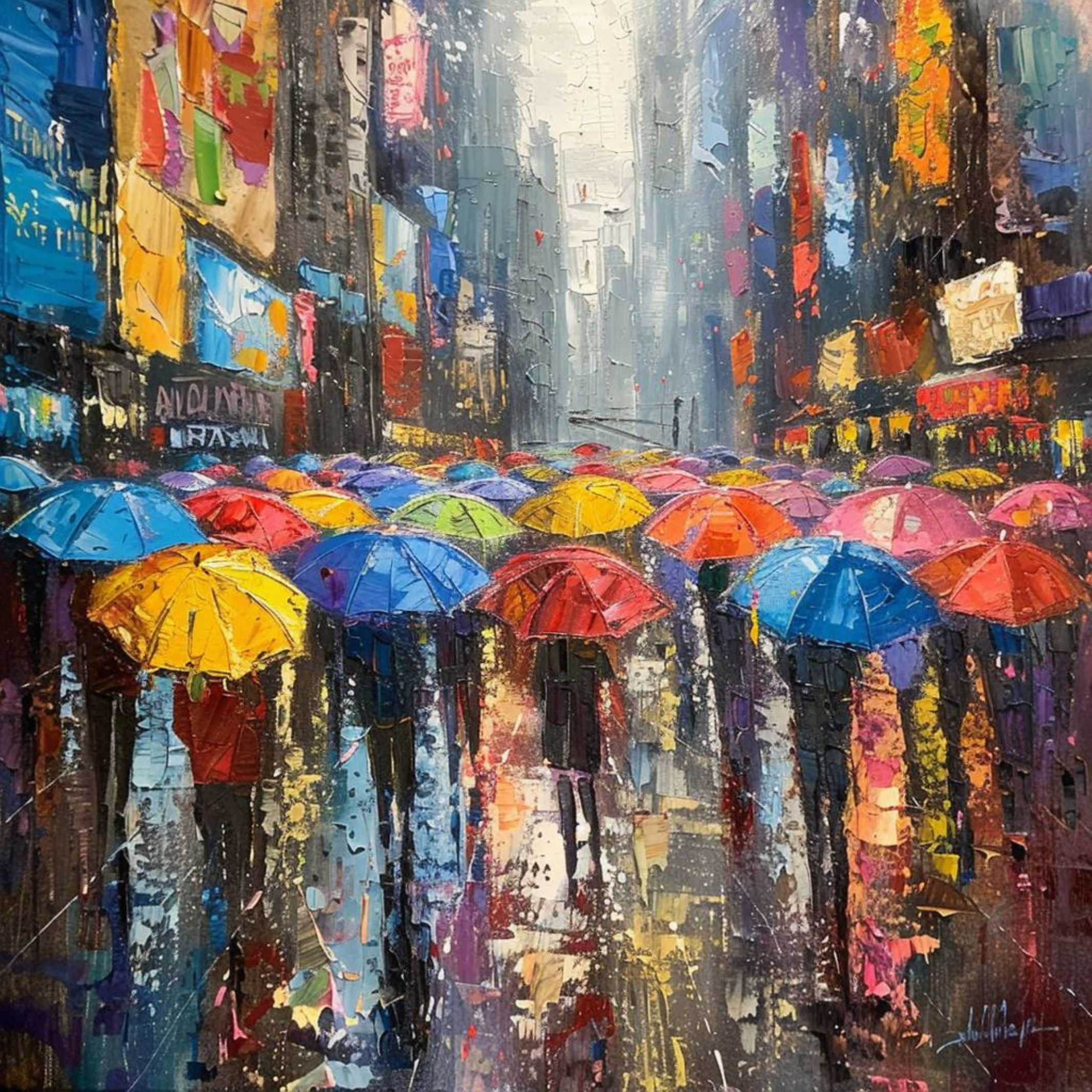} \\

\makecell[c]{GPT-\\Image-1} &
\img{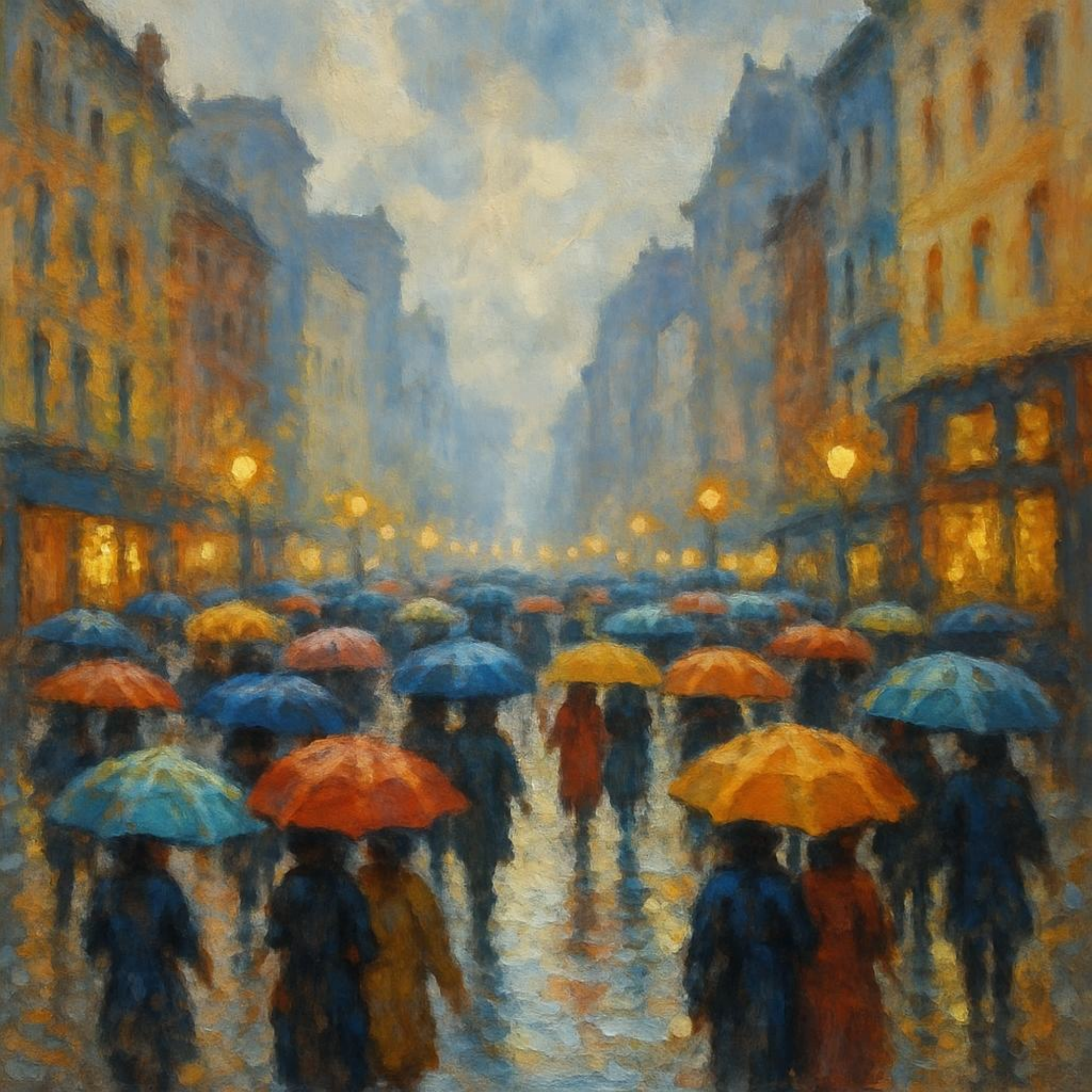} &
\img{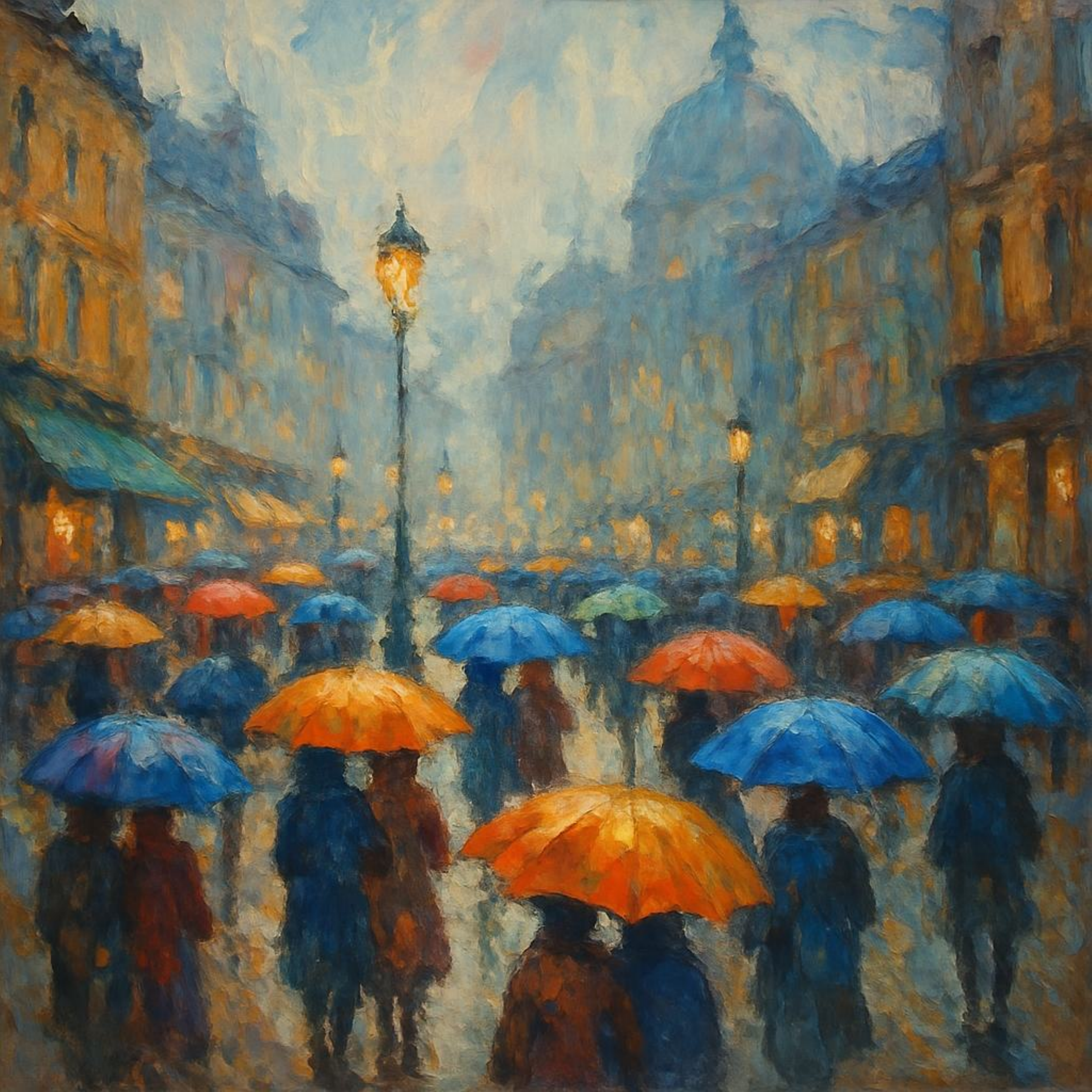} &
\img{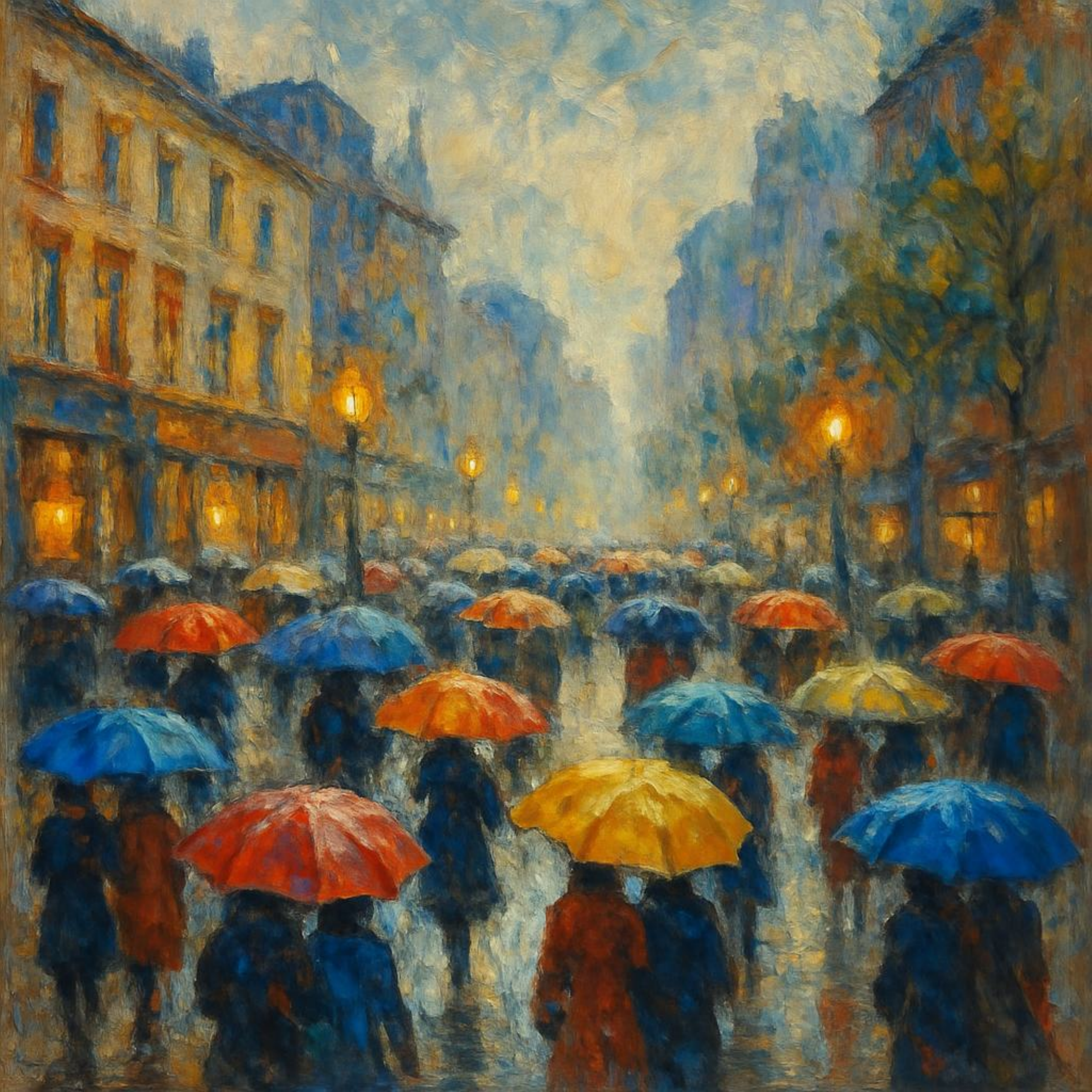} &
\img{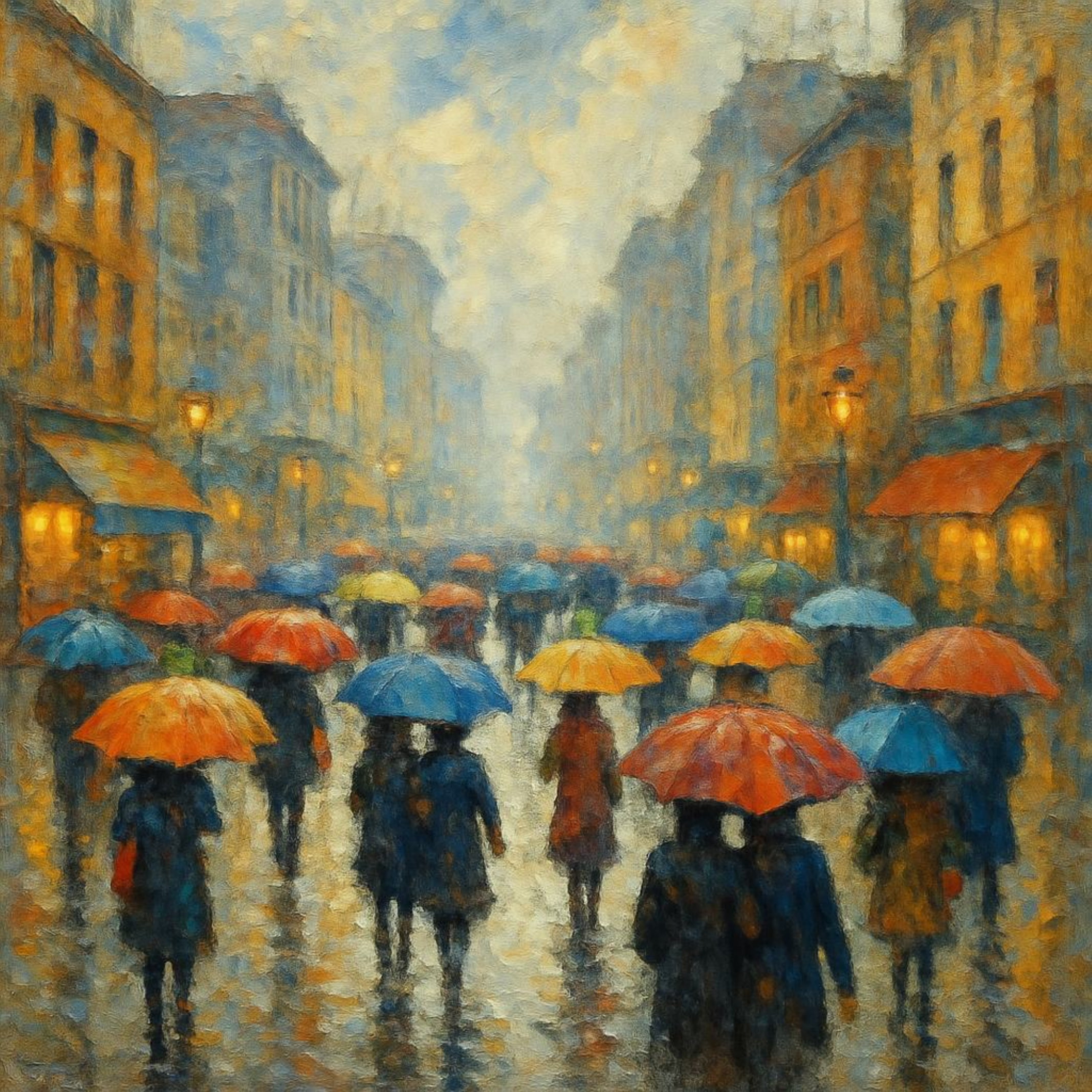} &
\img{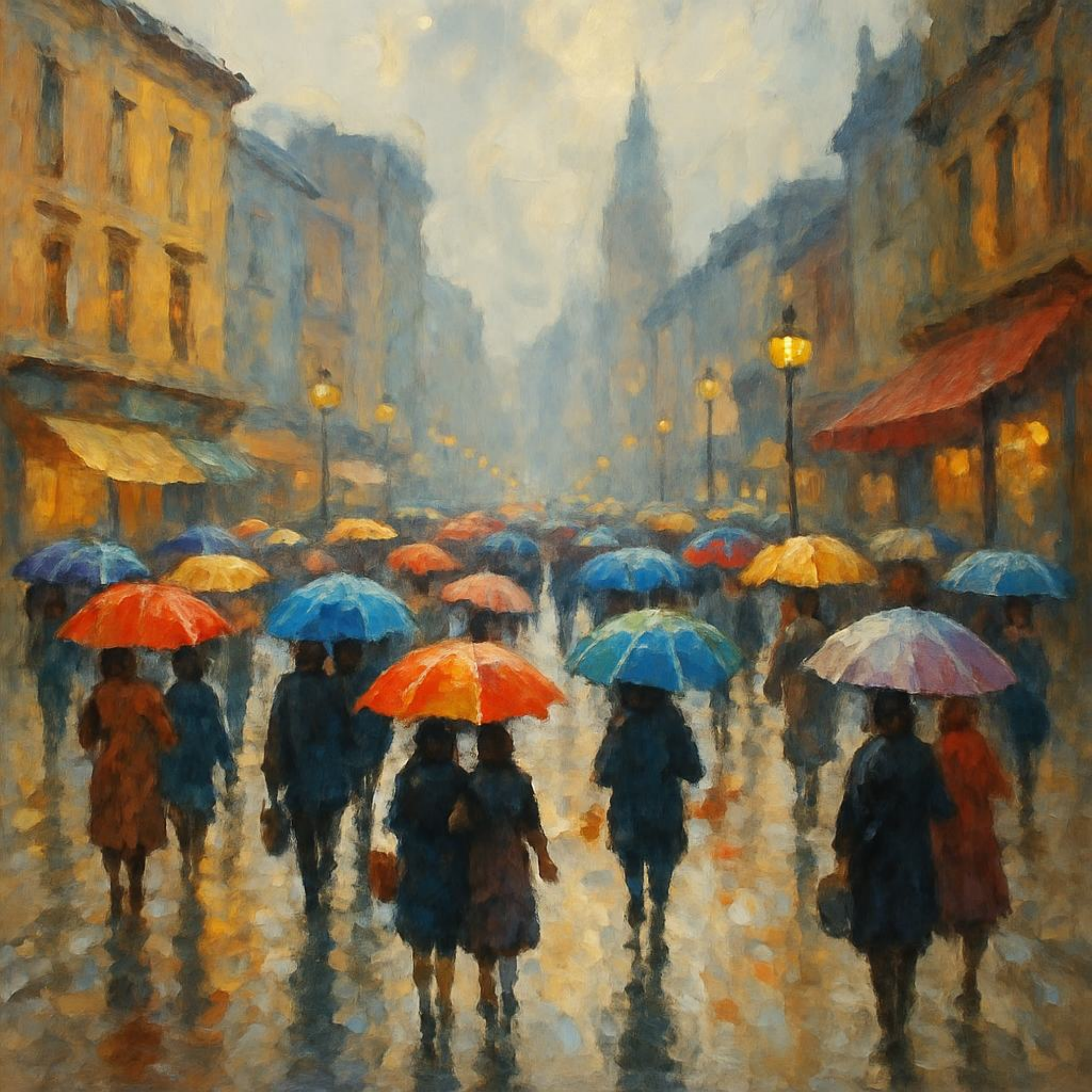} \\

\makecell[c]{Stable Diffusion\\3.5--large turbo} &
\img{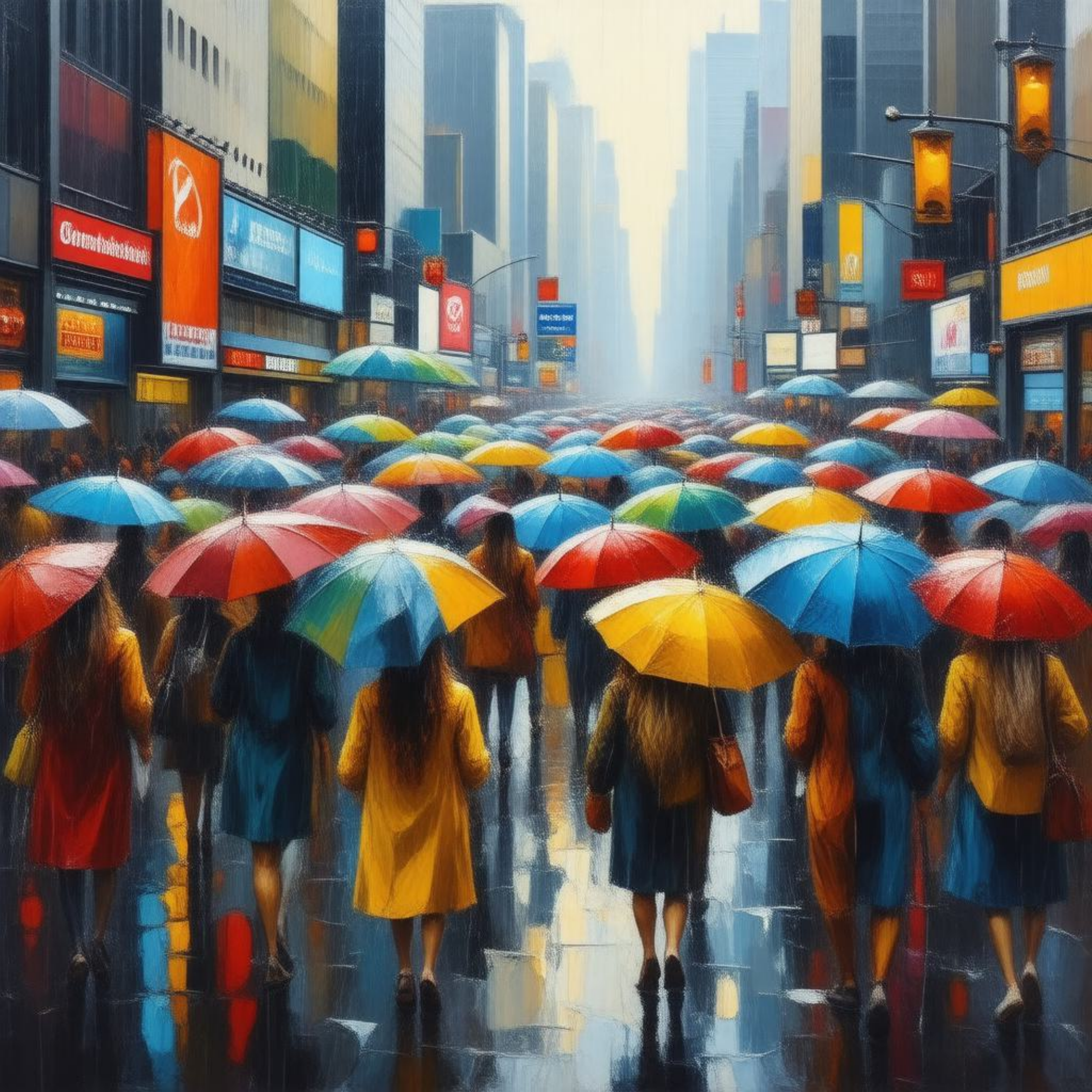} &
\img{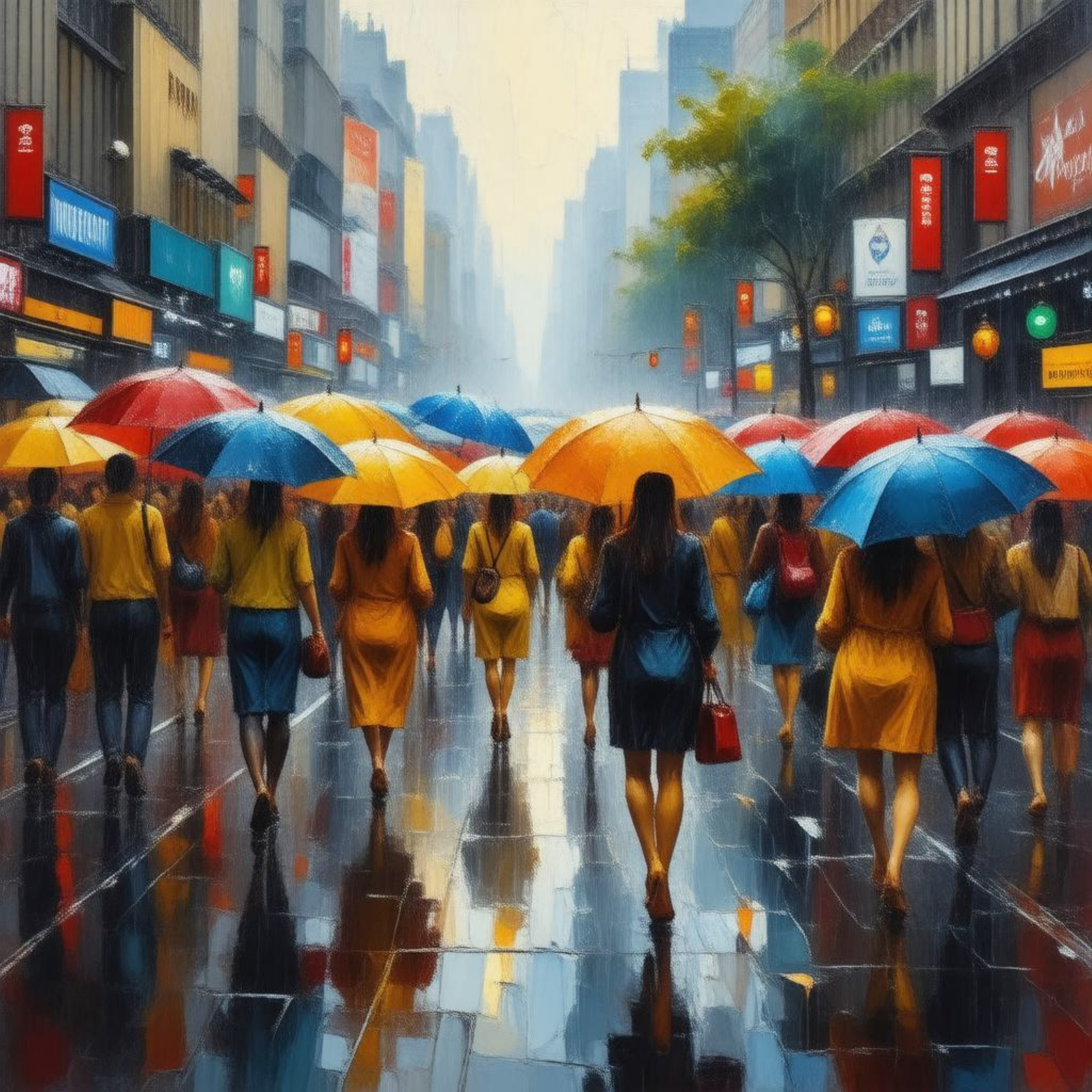} &
\img{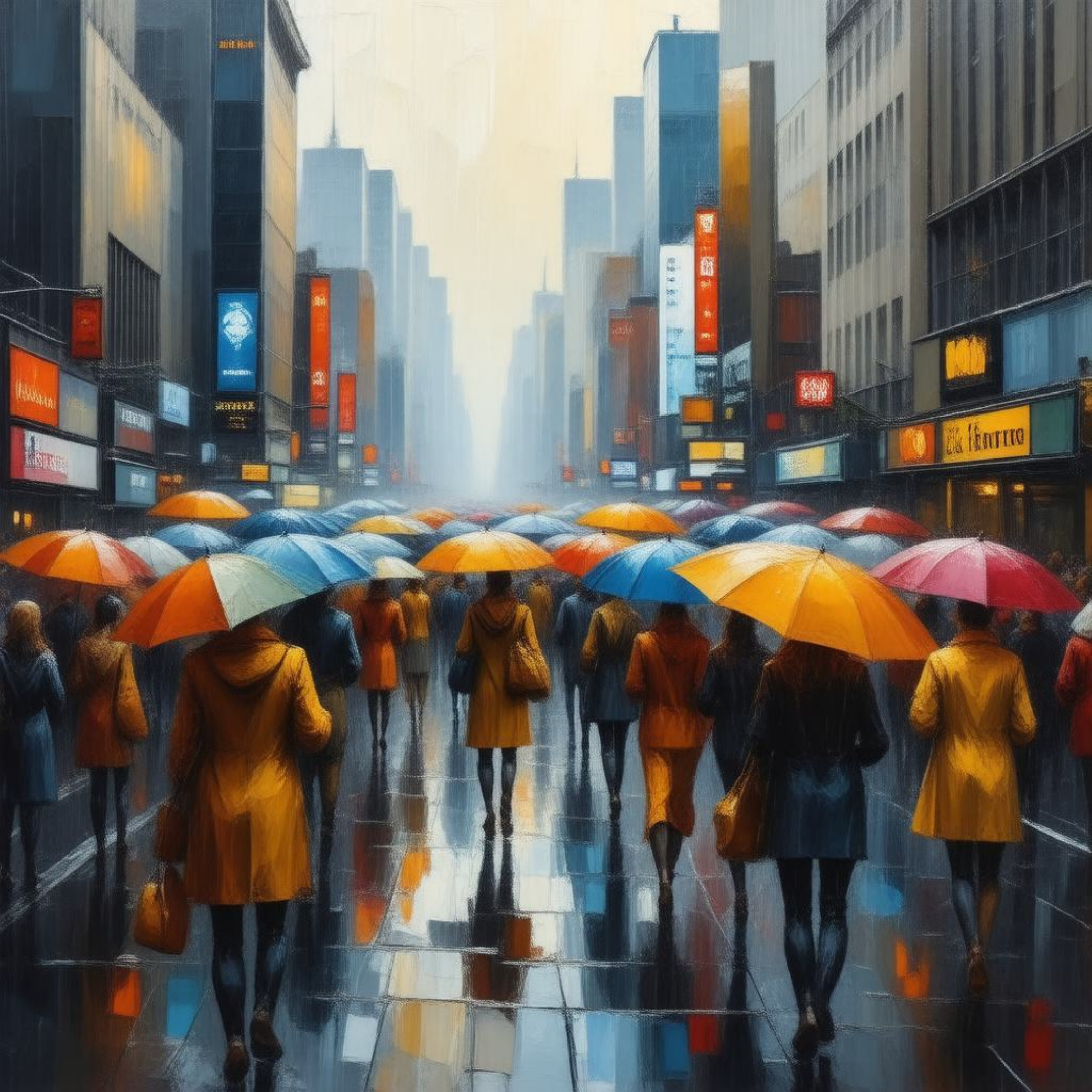} &
\img{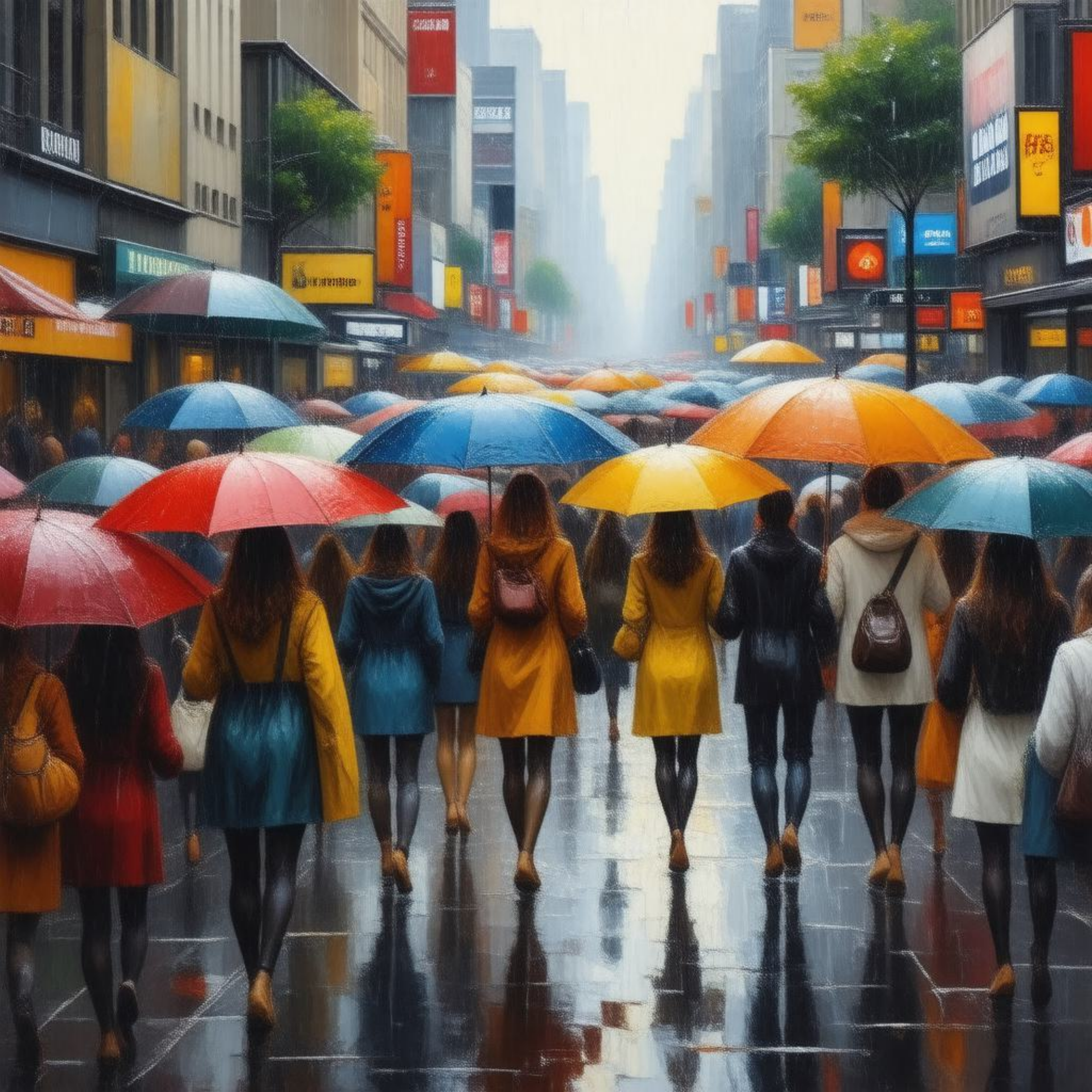} &
\img{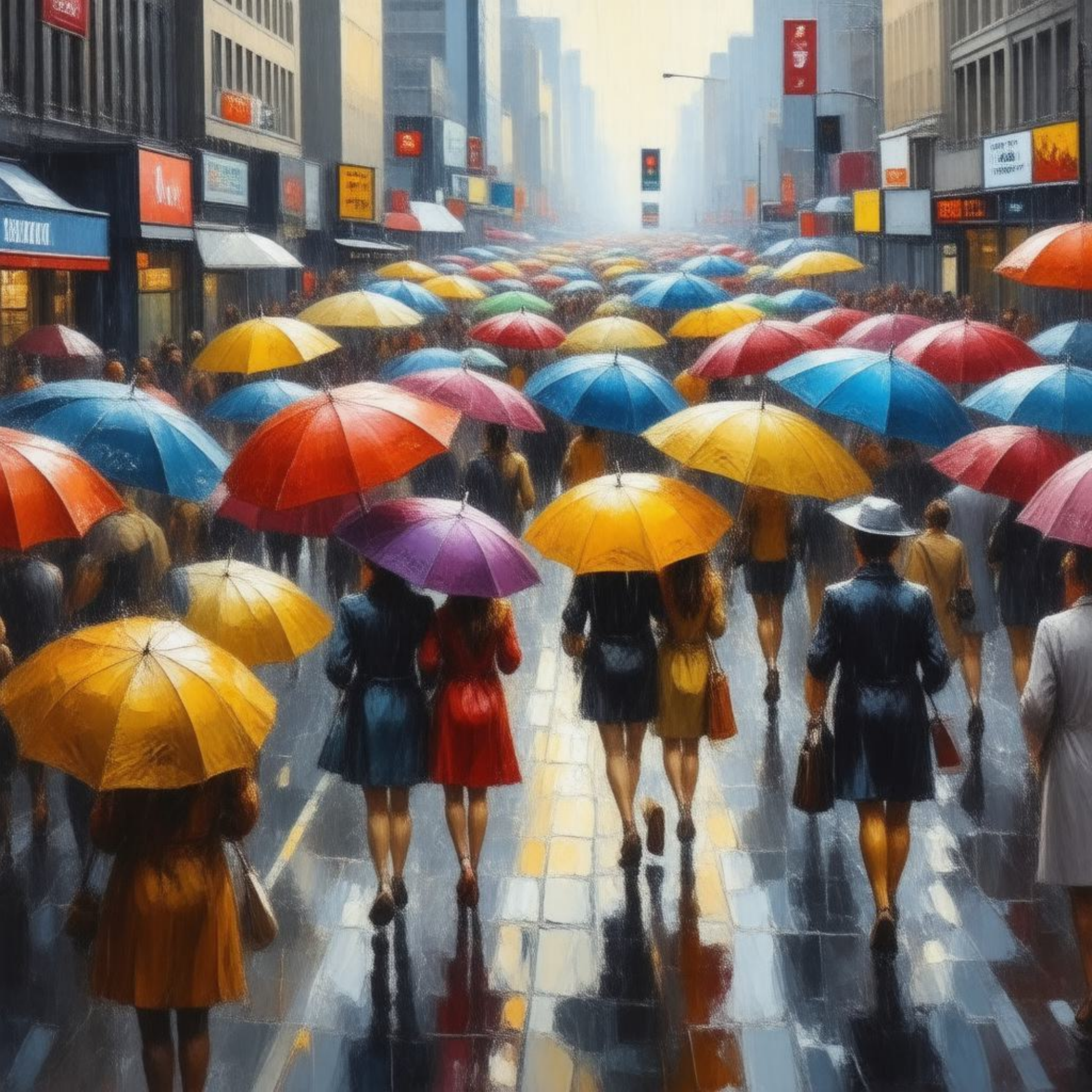} \\

\makecell[c]{HiDream-I1} &
\img{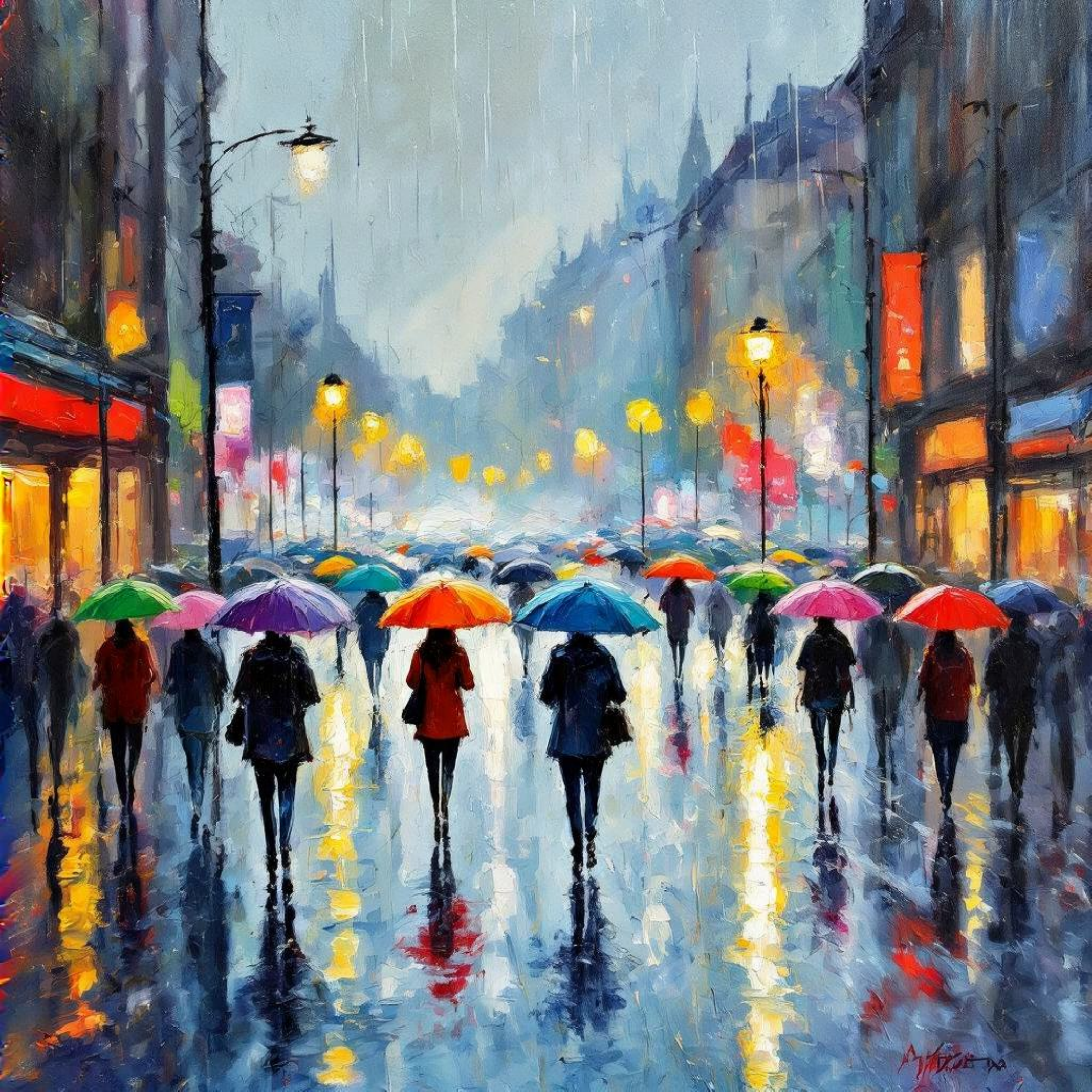} &
\img{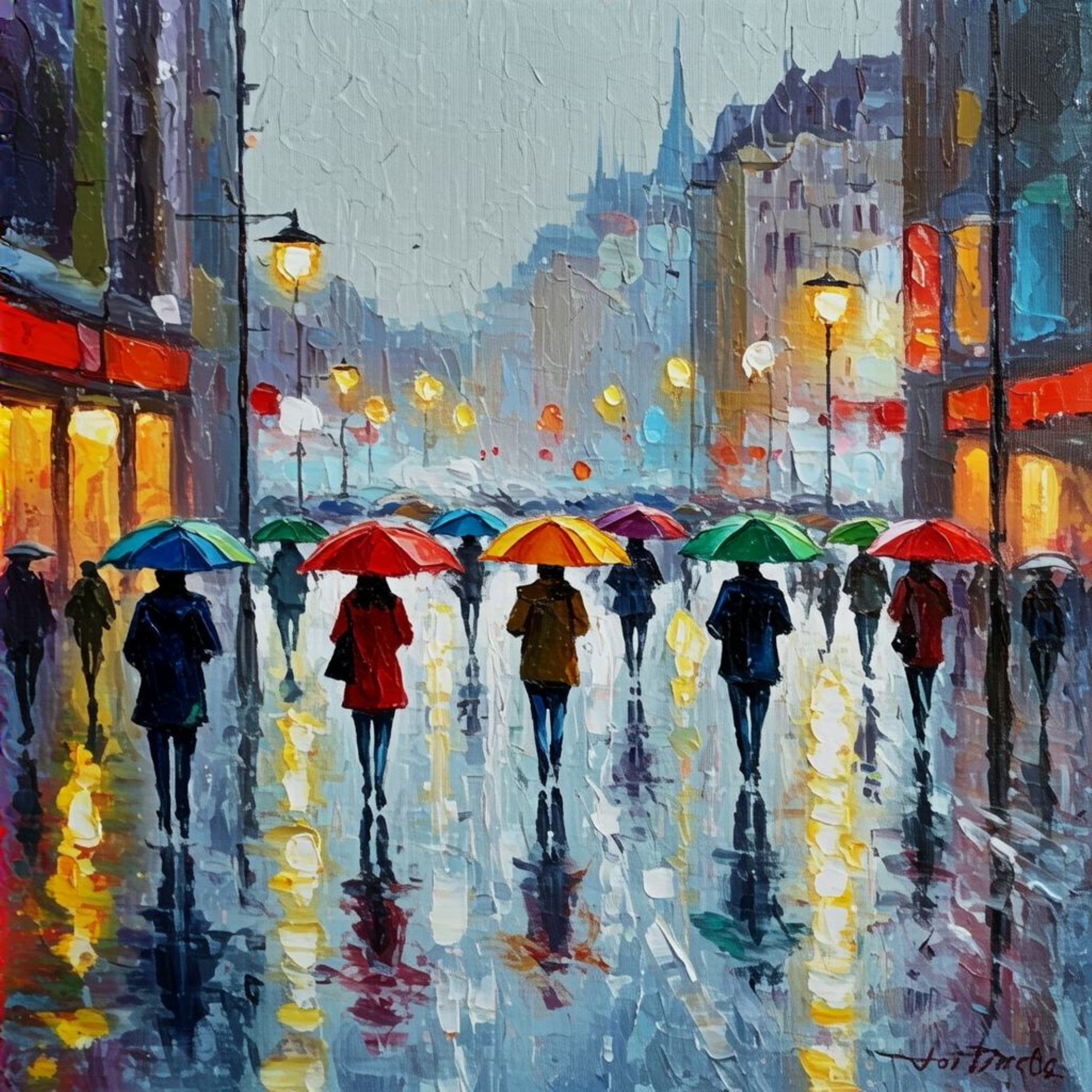} &
\img{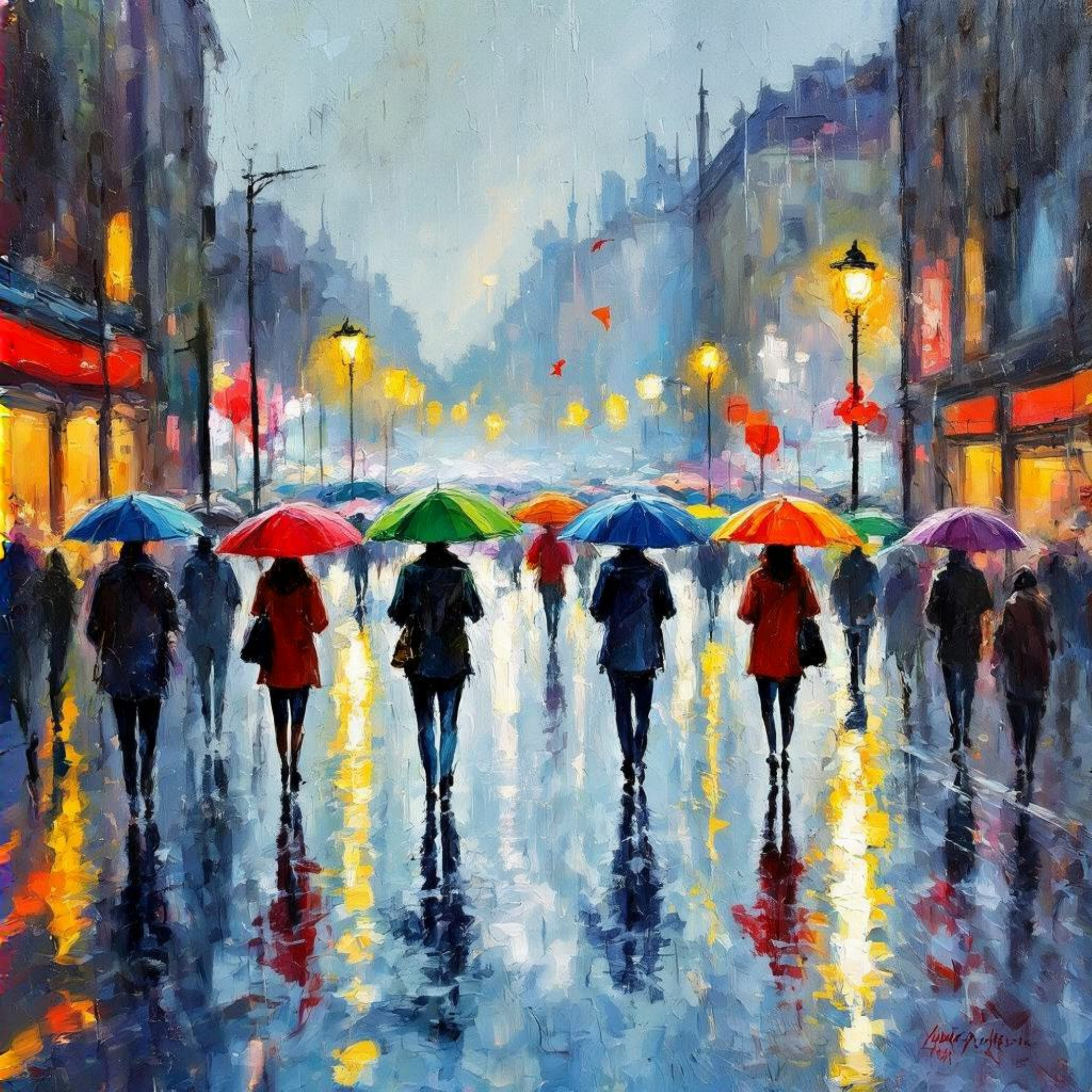} &
\img{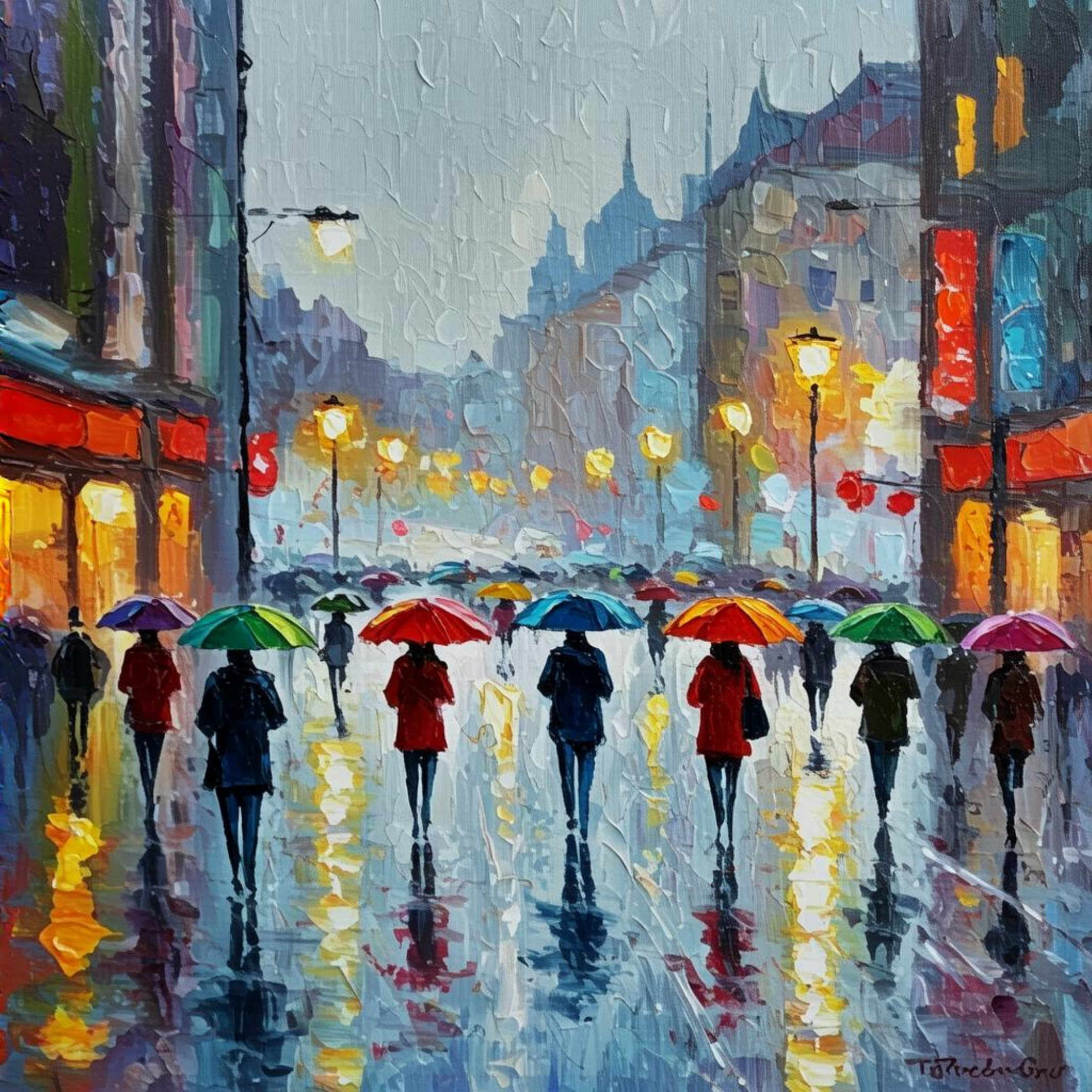} &
\img{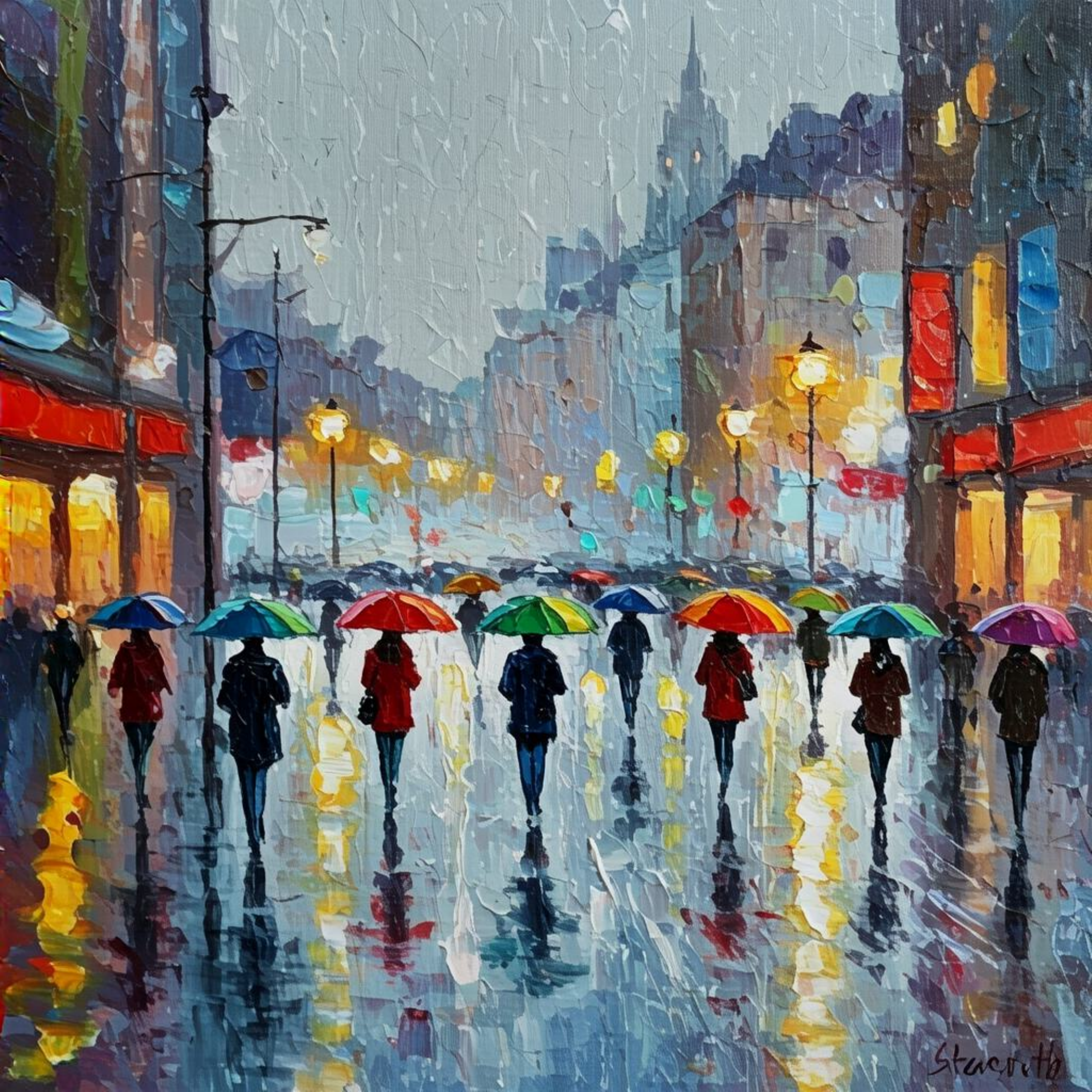} \\

\bottomrule
\end{tabular}%
} %

\caption{
Model-specific generation patterns for a fixed prompt.  
Each row shows five images from one model with different seeds, showing low intra-model diversity and strong inter-model differences.
}
\label{fig:prompt_diversity}
\end{figure}

%% file: sections/02_related_work.tex
\section{Related Work}
\subsection{Leaderboard Attacks}

Leaderboards for generative AI are generally either benchmark-based \cite{muennighoff2023mteb, minixhofer2024ttsds} or voting-based \cite{tts-arena-v2} (e.g., Chatbot Arena \cite{chiang2024chatbot}). Both types are vulnerable to manipulation attacks. \citet{huang2025exploring} demonstrate that malicious participants can deanonymize models in Chatbot Arena and artificially promote their own models through poisoned votes. \citet{zhao-etal-2025-challenges} show how inserting as little as 10\% adversarial/low-quality votes can shift a model’s rank by up to five places. \citet{min2025improving} analyze Elo-style rating systems, showing they can be gamed via ``omnipresent rigging," where a few hundred strategically placed votes can boost a model's rank substantially, even without targeting the victim directly. \citet{suri2026exploiting} examine leaderboards across several modalities and show that benchmark-based leaderboards can also be subverted by submitting models trained directly on test sets. Existing works mainly target LLMs or rely on backdoor-style deanonymization that may not generalize to T2I leaderboards.

\subsection{Model Attribution}

Work on model attribution seeks to infer the exact model given some form of access, typically through an API for querying. Prior work explored model attribution for GANs \cite{yu2019attributing} or focused on generative text modeling \cite{pasquini2025llmmap}.
Recent approaches focusing on T2I models either utilize adversarial examples for model attribution, requiring multiple API calls \cite{guo2024one}, or require tens of thousands of examples to train detection models, resulting in high false positive rates for unconstrained prompts \cite{yao2025authprint}.

%% file: sections/03_methodology.tex
\section{T2I De\-anonymization}

\subsection{Threat Model}
The adversary's objective is to deanonymize T2I models in order to manipulate their rankings on a voting-based leaderboard: by inferring which model a given anonymized generation corresponds to, the adversary can decide which model(s) to upvote or downvote.
We assume that the adversary has no control over the input prompts and aims to manipulate the ranking of \emph{any} model, not merely to identify its own.
For completeness, we also consider a stronger adversary that \emph{can} control the input prompts (Section \ref{subsec:prompt_control}) and find that under such conditions, deanonymization is even easier. 

\subsection{Methodology}
Generative models often produce characteristic outputs for the same input based on differences in training data, architecture, or even model size.  
These characteristics can be utilized as subtle ``signatures'' in the generated content.  
For T2I models, we hypothesize that the diversity of outputs from a given model across multiple generations of the same prompt is relatively low, while these outputs differ systematically from those of other models in style, content, or other features not explicitly described in the prompt.  
As these differences are largely semantic rather than pixel-level, we represent images in an embedding space that captures high-level features.  
Specifically, we employ CLIP \cite{radford2021learning} embeddings, which are well suited for semantic comparisons and can effectively highlight these model-specific generation patterns.

Our deanonymization algorithm, described in Algorithm \ref{alg:deanon}, proceeds as follows.
For each prompt $p$ from the leaderboard, we send it to every T2I model $M_i$ in a candidate set $\mathcal{C}$ and generate $k$ images per model. 
We embed both the leaderboard-provided image and all generated images into the CLIP space. 
We then compute the centroid $c_i$ of its $k$ embeddings. 
We compute distances from the embedding of the provided image to each centroid $c_i$ and sort models by these distances.
The model with the smallest distance is predicted to be the source of the leaderboard image.

\input{sections/alg_classification}

\subsection{Distinguishability Metric}
\label{sec:distinguishability}

To better understand the separability of different models' generations in the embedding space,  we introduce a metric that quantifies the \emph{distinguishability} of prompts. 
This metric helps identify prompts that yield highly separable clusters and thus enable stronger deanonymization.

\paragraph{Model-level Separability.}
For each prompt $p_i$ and each model $M_j$, we collect the $k$ embeddings of the images generated by $M_j$ on $p_i$, 
denoted $\{e_{i,j}^{(1)}, \dots, e_{i,j}^{(k)}\}$. 
For every embedding $e_{i,j}^{(\ell)}$, we find its nearest neighbor in the joint embedding set of all models for the same prompt.  
If the nearest neighbor also originates from $M_j$, we mark $e_{i,j}^{(\ell)}$ as correctly clustered.  
Let
\[
\text{frac}(i,j)
    = \frac{1}{k}\sum_{\ell=1}^{k}\mathbb{I}
      \bigl[\mathrm{NN}(e_{i,j}^{(\ell)}) \in M_j\bigr],
\]
where $\mathbb{I}[\cdot]$ is the indicator function and $\mathrm{NN}(\cdot)$ denotes the nearest neighbor.  
If $\text{frac}(i,j) > \tau$ for a chosen threshold $\tau \in (0,1)$, 
we call the cluster corresponding to $(i,M_j)$ \emph{separable}.

\paragraph{Prompt-level Distinguishability.}
The distinguishability score of prompt $p_i$ is then defined as
\[
D(i) = \frac{1}{|\mathcal{C}|}\sum_{M_j \in \mathcal{C}}\mathbb{I}[\text{frac}(i, j) > \tau],
\]
i.e., the fraction of models that form separable clusters under $p_i$.

A high value of $D(i)$ indicates that generations for prompt $p_i$ form well-separated clusters in the embedding space,
making it easier to deanonymize models based on that prompt.  
This metric thus provides a principled way to rank prompts by their power to reveal model identities.

%% file: sections/alg_classification.tex
\begin{algorithm}[t]
\caption{Centroid-based Deanonymization of T2I Models}
\label{alg:deanon}
\textbf{Input:}  Prompt $p$ from leaderboard,
         candidate models $\mathcal{C}=\{M_1,\dots,M_n\}$,
         number of samples $k$,
         leaderboard-provided image $I^\ast$,
         image encoder $\phi(\cdot)$ (e.g., CLIP)

\textbf{Output:}  Predicted generating model $\hat{M}$

\begin{algorithmic}[1]
\State $e^\ast \gets \phi(I^\ast)$ \Comment{Embed leaderboard-provided image $I^\ast$}

\For{each $M_i \in \mathcal{C}$}
  \State Generate $k$ images $\{I_{i,1},\dots,I_{i,k}\}$ with prompt $p$
  \State Compute embeddings $E_i = \{\phi(I_{i,1}),\dots,\phi(I_{i,k})\}$
  \State Compute centroid $c_i = \frac{1}{k}\sum_{j=1}^k E_{i,j}$
\EndFor

\State Compute distances $d_i = \| e^\ast - c_i \|_2$ for all $M_i \in \mathcal{C}$
\State $\hat{M} \gets \arg\min_{M_i \in \mathcal{C}} d_i$
\State \Return $\hat{M}$
\end{algorithmic}
\end{algorithm}

%% file: sections/04_experiments.tex
\section{Experiments}

\subsection{Settings}

\paragraph{Models and Dataset.}
We evaluate our method on a diverse set of 19 T2I models drawn from a broad spectrum of companies and organizations, including OpenAI, Midjourney, Stability~AI, HiDream.ai, Black Forest Labs, Playground~AI, Alibaba, and Alpha-VLLM.  
This collection spans multiple architectures within individual companies and includes multiple model sizes within the same architecture, providing diversity in both design and scale.
We evaluate using a set of 280 prompts collected from the ArtificialAnalysis\footnote{\url{https://artificialanalysis.ai/text-to-image/arena}} T2I leaderboard.  
A complete list of the models is provided in Table~\ref{tab:models}.

\paragraph{Hyperparameters and Evaluation Metric.}
We generate images at a resolution of $1024 \times 1024$ pixels, unless a model does not support it.  
CLIP embeddings are computed after resizing all images to $224 \times 224$ pixels (the standard CLIP input size) \emph{without cropping}, so differences in generation resolution have negligible effect on the final embeddings.  
For each prompt and model, we generate multiple images using different random seeds to capture intra-model variation.  
The number of inference steps for each model follows the default or recommended settings reported on the ArtificialAnalysis leaderboard website.  
To evaluate deanonymization performance, we compute and report \emph{top-k} accuracies (for $k\in[1,5]$), which measure the probability that the correct model appears within the corresponding number of top predictions.

\begin{figure}[t]
    \centering
    \includegraphics[width=0.80\linewidth]{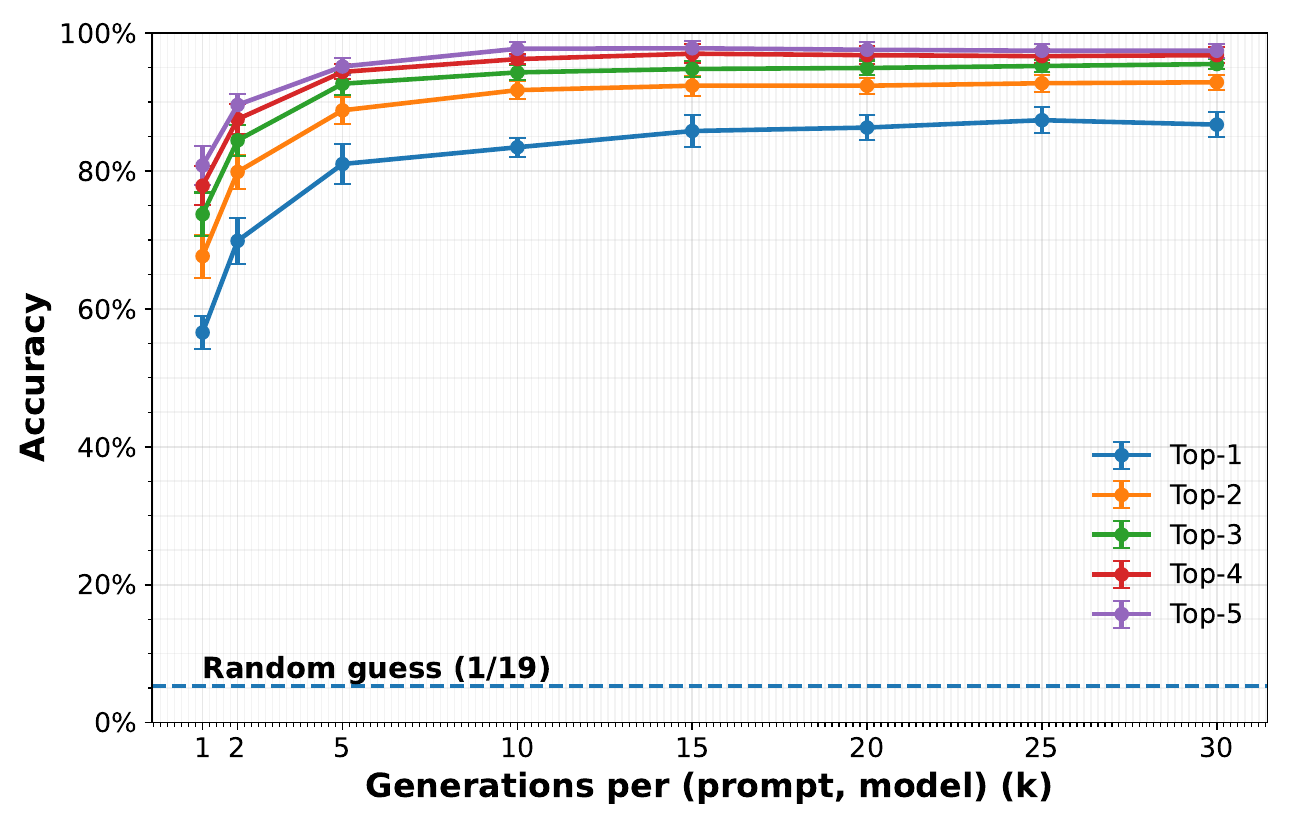}
    \caption{
Deanonymization accuracy versus number of generations $k$ per (prompt, model) pair.  
Curves show mean Top-1–Top-5 accuracy over five runs with one-standard-deviation error bars.  
The dashed line indicates the random-guess baseline of $1/19$.
}
    \label{fig:deanon_accuracy_vs_k}
\end{figure}

\subsection{Results}

\paragraph{Deanonymization Performance.}
Our centroid-based approach effectively predicts the model responsible for a given leaderboard image, achieving roughly $87\%$ \emph{top-1} accuracy, far exceeding the random-guess baseline of $\approx5.26\%$.  
Figure~\ref{fig:deanon_accuracy_vs_k} shows this advantage extends beyond the first prediction: \emph{top-3} accuracy reaches about $95\%$, meaning the correct model typically ranks among the very top candidates.

\paragraph{Effect of $k$.}
Figure~\ref{fig:deanon_accuracy_vs_k} also illustrates the influence of $k$, the number of generations per (prompt, model) pair used to compute centroids.  
Even a single generation ($k=1$) achieves \emph{top-1} accuracy of approximately $57\%$.  
Increasing $k$ substantially improves performance by providing more representative points in the embedding space, creating tighter and more robust model-specific clusters.  
We observe diminishing returns beyond 10-15 generations per prompt, where deanonymization accuracy saturates to near-perfect values.

\paragraph{Effect of Architecture and Model Size.}
To examine whether model architecture or size---including different architectures within the same company---impacts deanonymization, 
we include multiple architectures and model scales in our evaluation.  
For instance, from Stability~AI we evaluate both \texttt{stable\_diffusion\_2.1} and \texttt{stable\_diffusion\_3}, and from the Flux family we include both \texttt{flux\_1\_dev} and \texttt{flux\_1\_schnell}.  
We also compare different sizes within the same architecture, such as \texttt{stable\_diffusion\_3\_5\_large} versus \texttt{stable\_diffusion\_3\_5\_medium}.  
Even in these closely related cases, our method consistently achieves high distinguishability: the misclassification rate between the two \texttt{stable\_diffusion\_3.5} size variants is only about $3\%$, and between the two Flux variants is roughly $3.8\%$.  
The method works well even for distinguishing between models released by the same company or between different-size variants of the same architecture.

\begin{figure}[t]
    \centering
    \includegraphics[width=1.0\linewidth]{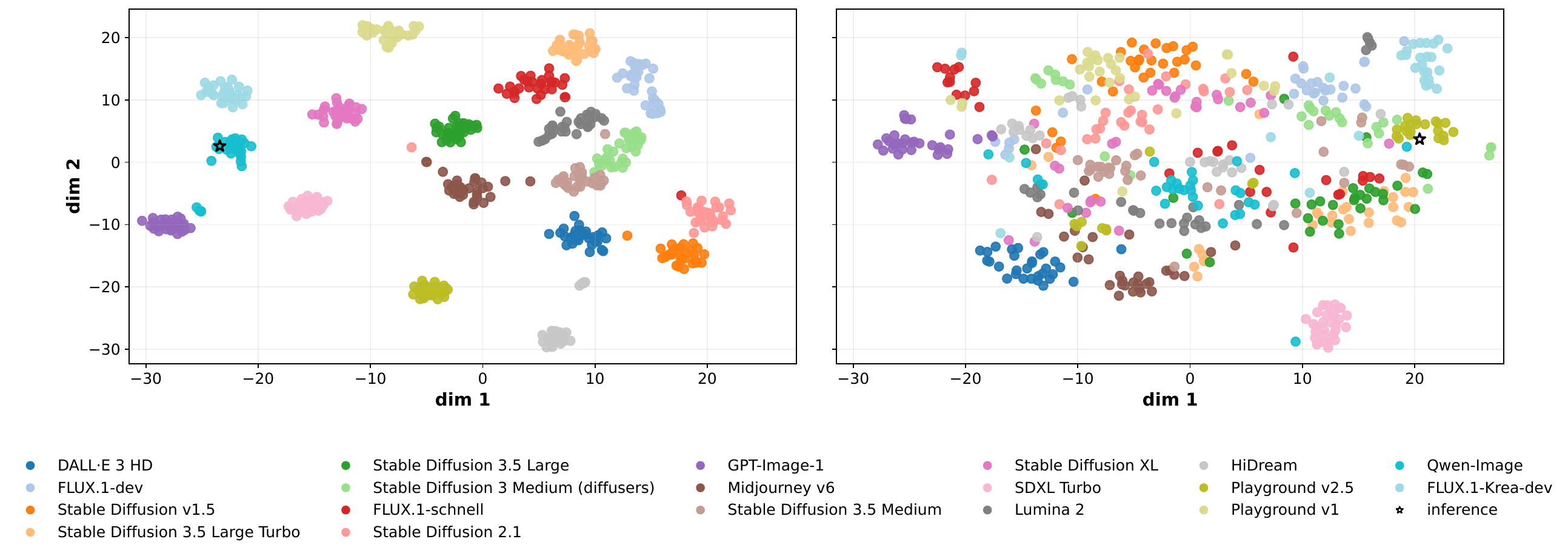}
    \caption{
CLIP-embedding visualizations for two representative prompts with contrasting distinguishability scores.  
\textbf{Left:} a high-distinguishability prompt (score $=1.0$), where generations from every model form clearly separated clusters.  
\textbf{Right:} a low-distinguishability prompt (score $=0.21$), where generations from different models overlap substantially, making deanonymization harder.
}
    \label{fig:dist_score_distribution}
\end{figure}

\paragraph{Distinguishability Score.}
From Section~\ref{sec:distinguishability}, the distinguishability score of a prompt is the fraction of models whose generations form separable clusters in the embedding space.  
Figure~\ref{fig:nn_dist} shows the distribution of this score across all 280 prompts. 
To illustrate the extremes, Figure~\ref{fig:dist_score_distribution} presents embedding visualizations for two representative prompts: one with a score of $1.0$, where generations from every model are perfectly separable, and another with a score of $0.21$, where most model clusters overlap substantially.
Higher distinguishability scores lead to higher deanonymization accuracy, confirming this metric as a strong predictor of attack success (Figure~\ref{fig:success_vs_dist}).

\subsection{Prompt-Controlled Attack}
\label{subsec:prompt_control}
Sorting the evaluation prompts by their distinguishability score reveals a small subset with a perfect score of $1.0$ (e.g., the left panel of Figure~\ref{fig:dist_score_distribution}), indicating that generations from all models form perfectly separable clusters.  
This observation implies that an adversary with the ability to craft their own prompts can achieve \emph{complete} distinguishability.
To evaluate this scenario, we randomly selected five prompts from our dataset with a distinguishability score of $1.0$.  
For each prompt we repeatedly (100 times) selected a random model, generated an image, and applied our centroid-based classification method to deanonymize it.  
This simple experiment achieves close to $99\%$ top-1 accuracy, confirming that an adversary with the ability to submit prompts to the leaderboard could reliably and confidently deanonymize the participating models.

\subsection{One-vs-Rest Classification}
Another practical scenario involves determining whether a given image was generated by a specific target model, rather than identifying the exact model among all candidates.  
This reduces to a \emph{one-vs-rest} classification problem in the CLIP embedding space.

For each evaluation prompt, we randomly select a model as the adversary’s target. We then generate an image using the given prompt and target model and compute the distances between the image’s embedding and (i) the centroid of the target model’s cluster and (ii) the centroids of all other model clusters. If the image embedding is closer to the target model’s centroid than to any other centroid, we classify it as generated by the target model; otherwise, we classify it as not. This simple approach achieves approximately $99\%$ accuracy.

We repeat this experiment by fixing the target model rather than selecting it randomly for each prompt.  
Specifically, for each of the 19 models we treat that model as the adversary’s target and evaluate on all 280 prompts, creating 19 fixed-target experiments.
Even for the model with the lowest performance, the prediction is correct in nearly $96\%$ of the cases, while two models—\texttt{HiDream} and \texttt{SDXL~Turbo}—reach perfect $100\%$  accuracy over all 280 samples (Table~\ref{tab:one_vs_rest_all_fixed}, Appendix~\ref{app:additional_results}).  
We also report AUC scores and TPRs at low FPRs for this setting; for example, \texttt{HiDream} and \texttt{SDXL~Turbo} both achieve TPRs of $1.0$ at FPR$=2\%$. An adversary targeting a specific model can abstain from voting when uncertain, effectively controlling their false positive rate.  
The TPR at a given FPR shows how many correct upvotes the adversary achieves while limiting false upvotes to other models; details of how we compute these AUC and TPR values are provided in Appendix~\ref{sec:auc_details}.

We also explore a more restrictive setup where the adversary has no access to other models (Appendix \ref{app:hardest_one_vs_rest}), finding that some models still achieve high distinguishability through outlier detection alone.

%% file: sections/05_conclusion.tex
\section{Conclusion}

In this work, we analyze text-to-image models and demonstrate how adversaries can successfully infer models based on generations in leaderboard arenas, despite lack of any control over the generation prompt.
This reveals a fundamental tension: the distinctive visual signatures that give models their competitive edge in quality and style are precisely what enable deanonymization attacks.
While recent work \cite{suri2026exploiting} suggests rotating prompts to prevent reuse, our results show this offers limited protection since models remain highly distinguishable even on unseen prompts.
More robust defenses might require analyzing voting patterns for anomalies or limiting the number of generations shown per prompt.
With growing concerns around the fairness and credibility in voting-based arenas for LLMs \cite{singh2025leaderboard}, understanding the extent of such deanonymization strategies is critical to actively design more secure leaderboards.

%% file: tables/model_lists.tex
\begin{table*}[ht]
\centering
\caption{
Full list of text-to-image models used in our experiments, along with their provider, image resolution, and the number of inference steps.
Where inference-step counts are available on the ArtificialAnalysis methodology page, we adopt those values directly.
For models not mentioned there, we use the default values documented on their respective Hugging Face model pages.  
For OpenAI and Midjourney models we did not explicitly set the number of inference steps and used their internal default generation settings.  
Finally, the model \texttt{Playground v1} does not expose an inference-steps parameter at all.
}
\small
\begin{tabular}{l l c c}
\toprule
\textbf{Model} & \textbf{Company / Provider} & \textbf{Resolution (W$\times$H)} & \textbf{Inference Steps} \\
\midrule
DALL·E 3 HD~\cite{betker2023improving} & OpenAI & 1024×1024 & -- \\
FLUX.1-dev~\cite{flux2024} & Black Forest Labs & 1024×1024 & 28 \\
Stable Diffusion v1.5~\cite{Rombach_2022_CVPR} & Stability AI & 512×512 & 50 \\
Stable Diffusion 3.5 Large Turbo & Stability AI & 1024×1024 & 4 \\
Stable Diffusion 3.5 Large & Stability AI & 1024×1024 & 35 \\
Stable Diffusion 3 Medium~\cite{esser2024scaling} & Stability AI & 1024×1024 & 30 \\
FLUX.1-schnell~\cite{flux2024} & Black Forest Labs & 1024×1024 & 4 \\
Stable Diffusion 2.1~\cite{Rombach_2022_CVPR} & Stability AI & 1024×1024 & 50 \\
GPT-Image-1 & OpenAI & 1024×1024 & -- \\
Midjourney v6~\cite{midjourney} & Midjourney & 1024×1024 & -- \\
Stable Diffusion 3.5 Medium & Stability AI & 1024×1024 & 40 \\
Stable Diffusion XL~\cite{podell2024sdxl} & Stability AI & 1024×1024 & 30 \\
SDXL Turbo~\cite{podell2024sdxl} & Stability AI & 1024×1024 & 4 \\
Lumina 2~\cite{qin2025lumina} & Alpha-VLLM & 1024×1024 & 50 \\
HiDream~\cite{cai2025hidream} & HiDream.ai & 1024×1024 & 50 \\
Playground v2.5~\cite{li2024playground} & Playground AI & 1024×1024 & 50 \\
Playground v1 & Playground AI & 1024×1024 & -- \\
Qwen-Image~\cite{wu2025qwenimagetechnicalreport} & Alibaba & 1024×1024 & 50 \\
FLUX.1-Krea-dev~\cite{flux2024} & Black Forest Labs & 1024×1024 & 28 \\
\bottomrule
\end{tabular}

\label{tab:models}
\end{table*}

%% file: tables/one_vs_rest_fixed_model.tex
\begin{table*}[h]
\centering
\caption{
Performance of the one-vs-rest deanonymization attack when each model is used as the fixed adversarial target across all 280 evaluation prompts.
The table reports per-model \emph{top-1 accuracy}, \emph{ROC--AUC}, and \emph{TPR} at two operating points (FPR$=2\%$ and FPR$=5\%$).
}
\label{tab:one_vs_rest_all_fixed}
\begin{tabular}{lcccc}
\toprule
\textbf{Model} & \textbf{Accuracy} & \textbf{ROC--AUC} & \textbf{TPR@2\%} & \textbf{TPR@5\%} \\
\midrule
HiDream & 1.000 & 1.000 & 1.000 & 1.000 \\
SDXL Turbo & 1.000 & 1.000 & 1.000 & 1.000 \\
DALL·E 3 HD & 0.996 & 0.998 & 1.000 & 1.000 \\
Playground v2.5 & 0.993 & 0.998 & 0.917 & 1.000 \\
FLUX.1-Krea-dev & 0.993 & 0.999 & 1.000 & 1.000 \\
Stable Diffusion 3.5 Medium & 0.989 & 0.995 & 0.929 & 0.929 \\
GPT-Image-1 & 0.989 & 0.992 & 0.938 & 0.938 \\
Stable Diffusion 3.5 Large & 0.986 & 0.998 & 1.000 & 1.000 \\
Stable Diffusion 3.5 Large Turbo & 0.986 & 0.998 & 0.929 & 1.000 \\
Stable Diffusion 3 Medium (diffusers) & 0.986 & 0.995 & 0.929 & 0.929 \\
FLUX.1-schnell & 0.982 & 0.990 & 0.846 & 0.923 \\
Qwen-Image & 0.979 & 0.955 & 0.733 & 0.733 \\
Stable Diffusion XL & 0.979 & 0.991 & 0.857 & 1.000 \\
Lumina 2 & 0.979 & 0.994 & 0.857 & 1.000 \\
Stable Diffusion 2.1 & 0.975 & 0.983 & 0.813 & 0.938 \\
Playground v1 & 0.975 & 0.978 & 0.944 & 0.944 \\
Midjourney v6 & 0.971 & 0.980 & 0.889 & 0.889 \\
Stable Diffusion v1.5 & 0.971 & 0.982 & 0.947 & 0.947 \\
FLUX.1-dev & 0.964 & 0.983 & 0.714 & 0.929 \\
\bottomrule
\end{tabular}
\end{table*}

%% file: tables/one_vs_rest_fixed_model_no_knowledge.tex
\begin{table*}[t]
\centering
\caption{
Results of the one-vs-rest attack when the adversary has access only to its target model’s generations.
We report per-model \emph{top-1 accuracy}, \emph{ROC--AUC}, and \emph{TPR} at two operating points (FPR$=2\%$ and FPR$=5\%$).
}
\label{tab:one_vs_rest_all_fixed_no_knowledge}
\begin{tabular}{lcccc}
\toprule
\textbf{Model} & \textbf{Accuracy} & \textbf{ROC--AUC} & \textbf{TPR@2\%} & \textbf{TPR@5\%} \\
\midrule
SDXL Turbo                                 & 0.993 & 0.996 & 1.000 & 1.000 \\
GPT-Image-1                                & 0.982 & 0.979 & 0.813 & 0.875 \\
HiDream                                    & 0.975 & 0.970 & 0.471 & 0.882 \\
Playground v2.5                            & 0.953 & 0.921 & 0.083 & 0.583 \\
DALL·E 3 HD                                & 0.939 & 0.945 & 0.250 & 0.563 \\
Stable Diffusion 3.5 Large Turbo           & 0.932 & 0.905 & 0.071 & 0.357 \\
Stable Diffusion 3.5 Large                 & 0.921 & 0.945 & 0.077 & 0.538 \\
FLUX.1-Krea-dev                            & 0.918 & 0.897 & 0.067 & 0.333 \\
Stable Diffusion 3 Medium (diffusers)      & 0.897 & 0.921 & 0.071 & 0.214 \\
Stable Diffusion 3.5 Medium                & 0.893 & 0.886 & 0.071 & 0.071 \\
FLUX.1-dev                                 & 0.879 & 0.825 & 0.286 & 0.357 \\
FLUX.1-schnell                             & 0.850 & 0.874 & 0.077 & 0.154 \\
Lumina 2                                   & 0.850 & 0.835 & 0.071 & 0.071 \\
Midjourney v6                              & 0.839 & 0.845 & 0.111 & 0.222 \\
Stable Diffusion XL                        & 0.829 & 0.785 & 0.000 & 0.143 \\
Playground v1                              & 0.821 & 0.828 & 0.111 & 0.111 \\
Stable Diffusion 2.1                       & 0.789 & 0.807 & 0.063 & 0.063 \\
Stable Diffusion v1.5                      & 0.747 & 0.798 & 0.053 & 0.053 \\
Qwen-Image                                 & 0.600 & 0.733 & 0.067 & 0.133 \\
\bottomrule
\end{tabular}
\end{table*}